\documentclass{article} 
\usepackage{float}
\usepackage{iclr2027_conference,times}

\usepackage{amsmath,amsfonts,bm}

\def\eqref#1{equation~\ref{#1}}

\def\1{\bm{1}}

\DeclareMathAlphabet{\mathsfit}{\encodingdefault}{\sfdefault}{m}{sl}
\SetMathAlphabet{\mathsfit}{bold}{\encodingdefault}{\sfdefault}{bx}{n}

\newcommand{\R}{\mathbb{R}}

\newcommand{\KL}{D_{\mathrm{KL}}}

\DeclareMathOperator*{\argmax}{arg\,max}
\DeclareMathOperator*{\argmin}{arg\,min}

\usepackage{hyperref}
\usepackage{url}

\usepackage{graphicx}
\usepackage{xspace}

\usepackage[utf8]{inputenc} 
\usepackage[T1]{fontenc}    
\usepackage{hyperref}       
\usepackage{url}            
\usepackage{booktabs}       
\usepackage{amsfonts}       
\usepackage{nicefrac}       
\usepackage{microtype}      
\usepackage{xcolor}         
\usepackage{svg}
\usepackage{enumitem}
\usepackage{booktabs,multirow,siunitx}
\usepackage{amsmath}
\usepackage{amssymb}
\usepackage{mathtools}
\usepackage{amsthm}
\usepackage{algorithm}
\usepackage{algorithmic}
\usepackage{bbm}
\usepackage{subcaption}
\newtheorem{assumption}{Assumption}

\newtheorem{proposition}{Proposition}
\newtheorem{lemma}{Lemma}
\newtheorem{theorem}{Theorem}
\newtheorem{corollary}{Corollary}
\newtheorem{remark}{Remark}

\newcommand{\JS}{\operatorname{JS}}
\newcommand{\diag}{\operatorname{Diag}}
\newcommand{\one}{\mathbbm{1}}
\newcommand{\calF}{\mathcal{L}_{\max}}
\newcommand{\calR}{\mathcal{L}_{\min}}
\newcommand{\calL}{\mathcal L}
\newcommand{\calD}{\mathcal D}
\newcommand{\calP}{\mathcal P}
\newcommand{\method}{\texttt{$f$-Fum}\xspace}

\title{ \method: Federated Unlearning via min--max and $f$-divergence}

\author{%
  Radmehr Karimian\thanks{Equal contribution.}\\
  Universite de Geneve\\
  Geneva, Switzerland \\
  \texttt{radmehr.karimian@etu.unige.ch} \\
  \And
  Amirhossein Bagheri\footnotemark[1] \\
  Politecnico di Milano\\
  Milan, Italy \\
  \texttt{amirhossein.bagheri@mail.polimi.it} \\
  \AND
  Meghdad Kurmanji \\
  University of Cambridge\\
  Cambridge, United Kingdom \\
  \texttt{mk2296@cam.ac.uk}
  \And
  Nicholas D. Lane \\
  University of Cambridge\\
  Cambridge, United Kingdom \\
  \texttt{ndl32@cam.ac.uk}
  \And
  Gholamali Aminian \\
  The Alan Turing Institute\\
  London, UK \\
  \texttt{gaminian@turing.ac.uk} \\
}

\iclrfinalcopy 
\begin{document}

\maketitle
\lhead{}

\begin{abstract}
Federated learning (FL) enables collaborative training while keeping raw
data on clients. Deletion requests and the discovery of poisoned data
create a need to remove selected contributions from an already trained
model. This is challenging in FL because client data are decentralized
and their influence is entangled through repeated aggregation. We
present \method, an active federated unlearning method that builds on
teacher-student forget/retain optimization. Starting from a pretrained
global model, the method keeps that model fixed as a teacher. Clients
maximize an $f$-divergence between student and teacher predictive
distributions on forget examples, then minimize KL-based
teacher-student disagreement and supervised loss on retained examples.
The updates are aggregated in separate, sample-weighted synchronization
phases without moving raw data to the server or changing the model
architecture. We evaluate forget-side divergence choices in client-level
and data-level deletion settings involving backdoors, label confusion,
and clean-data deletion. Under equal synchronization-phase budgets,
divergence choice changes the forgetting-utility trade-off, yielding
improvements in some settings and mixed results in others. In the
evaluated configurations, \method uses up to $8\times$ fewer
synchronization phases than full federated retraining.
\end{abstract}

\section{Introduction}

The rapid progress of modern machine learning (ML) systems is largely driven by the abundance of data. However, the centralization of vast datasets raises critical privacy and security concerns, especially in sensitive domains such as healthcare, finance, and mobile computing. To mitigate these risks, Federated Learning (FL) has emerged as a powerful collaborative training paradigm. In FL, multiple clients jointly train a global model while keeping their raw data local. By sharing only model updates rather than the data itself, FL fundamentally changes the landscape of privacy-preserving ML and AI systems.

Although FL reduces the need to centralize private data, it does not inherently address what should happen after training when data must be removed. Such removal may be required due to user deletion requests enforcing the European Union’s Right to Be Forgotten~\citep{european_union_gdpr_2016}, client withdrawal, or the discovery of poisoned and mislabeled data. These concerns have motivated the study of machine unlearning (MU), which aims to remove the influence of specific training data from an already trained model~\citep{CaoYang2015}. The most direct solution, known as exact unlearning, requires retraining the model from scratch without the targeted data. However, exact retraining is computationally prohibitive for deep models and is especially impractical in FL, where data is distributed and communication rounds are costly. To bypass these costs, approximate unlearning methods seek to modify an existing model to behave similarly to a retrained model, substantially reducing overhead. In the federated setting, this gives rise to federated unlearning (FU), where the objective is to efficiently remove the influence of selected samples, updates, or entire clients from the global model while preserving performance on retained data.

Achieving effective FU is fundamentally more complex than traditional centralized MU. In centralized ML, the model owner has direct access to both the full training dataset and the model, making unlearning straightforward. In contrast, FL operates under a decentralized approach where data remains strictly on clients' devices. Consequently, methods designed for centralized unlearning usually cannot be transferred directly to FL. Furthermore, the collaborative nature of FL deeply integrates individual client updates into the global model, making it highly challenging to isolate and remove a single participant's effect without access to the original raw data. FU also necessitates a broader operational scope: rather than just deleting individual data points, it often requires the elimination of entire clients or groups of historical updates. Thus, FU must operate with severely limited information while strictly adhering to the communication constraints and privacy guarantees inherent to the federated ecosystem.

In practice, the success of unlearning also depends strongly on data heterogeneity, model architecture, federation size, and the severity of the unlearning request. Different unlearning scenarios vary significantly, ranging from benign client withdrawal (clean-data deletion) to the removal of adversarial contributions like poisoned updates or backdoors. Navigating these diverse and complex conditions requires a robust unlearning framework that can systematically isolate targeted information without causing a catastrophic collapse of the global model's utility.



Our method builds on the teacher-student forget/retain principle proposed by \citep{kurmanji2023unboundedmachineunlearning}. We study its use in active federated unlearning, where forget and retain examples remain at clients and the two updates use separate, sample-weighted aggregation phases. Our formulation permits different $f$-divergences for the retain and forget objectives. We analyze the local curvature of this objective under stated assumptions and evaluate how divergence choice affects the forgetting utility trade-off under a fixed synchronization-phase budget. Our main contributions are as follows:
\begin{itemize}
    \item Inspired by $f$-divergences, we proposed the forget/retain objective for
    active federated unlearning through client-local updates and
    standard sample-weighted aggregation, supporting both client-level
    and data-level deletion requests without server access to raw data. Our algorithm avoids catastrophic collapse and provides a stable unlearning process.

    \item We study the choice of $f$-divergence for forgetting while
    retaining KL-based alignment, and derive a local curvature
    characterization that makes the role of the chosen divergence
    explicit. We show that the choice of divergence has a substantial effect on optimization stability, retained-data utility, and forgetting behavior.

    \item We evaluate our federated unlearning method across backdoor removal,
    label-confusion removal, and privacy-oriented deletion on several
    datasets and federation sizes. We report retained utility together
    with forgetting metrics and identify settings in which the method
    performs well or exhibits trade-offs.
\end{itemize}
 \vspace{-1em}
\section{Related Work}

In this section, we will provide an overview of federated unlearning works focusing on \emph{active FU}. More related works are provided in Appendix \ref{appendix:A}.


Federated unlearning (FU) can broadly be divided into two categories: \textit{active unlearning} and \textit{passive unlearning}. In the active setting, the client requesting data deletion actively engages in the unlearning process, helping the system mitigate its own contribution to the global model. On the other hand, passive unlearning assumes that the client wishing to be forgotten is either unavailable or unwilling to participate. (e.g. \citep{liu2021revfrf}, \citep{su2023asynchronous}, \citep{li2023subspace}, \citep{cao2023fedrecover})

\paragraph{Passive Federated Unlearning:} Most passive unlearning approaches attempt to reconstruct the model without access to the forgetting client (e.g., \citep{liu2021federaser}, \citep{jiang2024efficientcertifiedrecoverypoisoning}, \citep{huynh2024fastfedultrainingfreefederatedunlearning}, \citep{yuan2023federated}). However, these methods have fundamental limitations. Primarily, they focus on removing an entire client, which may not be practical in many scenarios. Additionally, they rely on estimating the gradients of the removed client and subtracting them from the final model, which can be error-prone and computationally expensive. 

\paragraph{Active Federated Unlearning:} We divide active unlearning into two categories: (1) settings in which the server acts purely as an \emph{accumulator} (i.e., it only aggregates client updates), and (2) settings in which the server performs additional computation beyond averaging e.g., leveraging a server dataset or replay memory, maintaining auxiliary state, or applying extra procedures such as special operators.

\paragraph{Only Accumulator:} \citep{halimi2023federatedunlearningefficientlyerase} proposed Federated Unlearning, where a client locally reverses its contribution by maximizing empirical loss under constraints, followed by limited retraining across remaining clients. \citep{khalil2025notfederatedunlearningweight} proposes perturbing the model’s weights, strong enough to push the model away from its current solution yet resilient enough to allow rapid recovery, followed by brief fine-tuning to restore performance.

\citep{dhasade2024quickdropefficientfederatedunlearning} integrates dataset distillation into the unlearning process, allowing clients to generate compact representations used for gradient ascent unlearning. 
\citep{10682764} combines confusion based updates and salience aware masking to weaken model memory of 
specific data (like \citep{ma2022learn}) By simulating memory degradation and avoiding full retraining, it enables lightweight, instance to client level unlearning in federated settings.

 \citep{li2023federated} proposes a novel Federated Active Forgetting (FedAF) framework that leverages teacher-student learning to generate new memories for overwriting old ones, addressing fine-grained unlearning challenges while mitigating catastrophic forgetting in FL settings. The \citep{gu2024unlearning} introduces FedAU, a streamlined Federated Machine Unlearning framework that integrates a lightweight auxiliary unlearning module during the training process, using a linear operation to enable efficient unlearning at sample, class, and client levels without additional time-consuming steps. You can find beyond accumulator literature in Appendix \ref{appendix:A}

 \paragraph{Positioning of our work:}
Our work belongs to the active and aggregator-only FU setting where the server is not assumed to hold auxiliary data, perform replay-based recovery, or apply architecture specific operations beyond aggregation. Unlike methods based primarily on direct weight perturbation, gradient reversal, or auxiliary unlearning modules,  \method formulates unlearning as a min-max optimization problem using $f$-divergence objectives.

\section{Preliminaries}


\textbf{$f$-\textbf{divergence}:} The $f$-divergences are information measures that generalize various divergences, such as Kullback-Leibler (KL) divergence, through the use of a convex generator function $f$. Given two discrete distributions $P = \{p_i\}_{i=1}^{k}$ and $Q = \{q_i\}_{i=1}^{k}$, the $f$-divergence between them is defined as:
\begin{equation}
D_f(P \parallel Q) := \sum_{i=1}^{k} q_i f\left( \frac{p_i}{q_i} \right)
    \label{eq:f-div}
\end{equation}
where $f: (0, \infty) \to \mathbb{R}$ is a convex function with the property that $f(1) = 0$. In this work, we focus on KL-divergence, JS-divergence and $\chi^2$-divergence. Therefore, we consider $D_f\in\{D_{\mathrm{KL}},D_{\mathrm{JS}}, D_{\mathrm{\chi^2}}\}$. The extra explanation and rationale behind this selection for JS and $\chi^2$ divergences is provided in Appendix~\ref{app:motive-js-chi}.

\paragraph{Federated unlearning setup:}
Let $C$ denote the number of clients and
$S=\bigcup_{c=1}^C S^{(c)}$ the distributed training set.
For client $c$, an unlearning request specifies
$S_F^{(c)}\subseteq S^{(c)}$, with retained set
$S_R^{(c)}=S^{(c)}\setminus S_F^{(c)}$. We write
\begin{equation}
    S_F=\bigcup_{c=1}^C S_F^{(c)},
    \qquad
    S_R=\bigcup_{c=1}^C S_R^{(c)}.
\end{equation}
A federated learning algorithm $\mathcal A$ produces the original model
$\theta_0=\mathcal A(S)$, while an unlearning algorithm $\mathcal U$ produces
\begin{equation}
    \theta_U=\mathcal U(\theta_0,S_F).
\end{equation}
The retained-data retraining reference is
$\theta_R=\mathcal A(S_R)$.

We use predictive discrepancy
$\Delta_{\mathrm{pred}}(\theta_U,\theta_R)$ to characterize approximate
unlearning:
\begin{equation}
    \Delta_{\mathrm{pred}}(\theta_U,\theta_R)
    \leq \epsilon_{\mathrm{unl}},
    \label{eq:approx-unlearning}
\end{equation}
where smaller $\epsilon_{\mathrm{unl}}$ indicates closer behavior to
retained-data retraining.

\textbf{Loss Functions:} Let $h(x;\theta)\in\Delta^{K-1}$ denote the predictive distribution.
For student and teacher parameters $\theta^S$ and $\theta^T$, respectively,
we define
\begin{equation}
    d_f(x;\theta^S,\theta^T)
    =
    D_f\!\left(
        h(x;\theta^S)\|
        h(x;\theta^T)
    \right),
    \label{eq:f-div-st}
\end{equation}
and use
$\ell_f(h(x;\theta),\mathbf Y)
=D_f(\mathbf Y\|h(x;\theta))$
as the supervised loss.

\paragraph{Unlearning objectives:}
We consider two evaluation regimes.
In \emph{robustness-oriented unlearning}, $S_F$ contains corrupted or
poisoned samples, and effectiveness is measured by removal of the attack
effect while preserving clean predictive utility.
In \emph{privacy-oriented unlearning}, $S_F$ contains samples requested for
deletion, and residual information is evaluated using membership-inference
attacks (MIAs).

For an attack $\hat m$ on the unlearned model,
\begin{equation}
\mathrm{MIA}(\theta_U)
=
\frac{1}{2}
\left(
\Pr_{x\sim S_F}[\hat m(\theta_U,x)=1]
+
\Pr_{x'\sim S_{\mathrm{test}}}[\hat m(\theta_U,x')=0]
\right),
\end{equation}
with corresponding advantage
\begin{equation}
    \operatorname{Adv}_{\mathrm{MIA}}(\theta_U)
    =
    \left|2\,\mathrm{MIA}(\theta_U)-1\right|.
    \label{adv}
\end{equation}
Values near zero indicate attack performance close to random guessing.
The specific MIA variants used in our experiments are described in
Section~\ref{sec:experiments}.

\section{\texorpdfstring{$\method \:$}{method} Algorithm}
In this section, we introduce our FU algorithm,  \method. In the  \method  setting, a client may request unlearning due to poisoned data or privacy concerns, after which all clients and the server collaboratively perform the unlearning process, i.e., active unlearning. Our algorithm consists of three main steps: (i) Local Objectives, (ii) Federated Objectives, and (iii) Post-training Minimization.
\method performs \(T\) max–min iterations, each comprising one forget-set maximization phase and one retained-data minimization phase, with a server aggregation after each phase. These iterations are followed by \(T_{\mathrm{min}}\) minimization-only phases, giving \(2T+T_{\mathrm{min}}\) synchronization phases in total.
\begin{figure*}[t]
    \centering

    \begin{subfigure}[t]{0.47\textwidth}
        \centering
        \vspace{0pt}
        \includegraphics[
            height=0.245\textheight,
            keepaspectratio,
            trim=5 5 5 5,
            clip
        ]{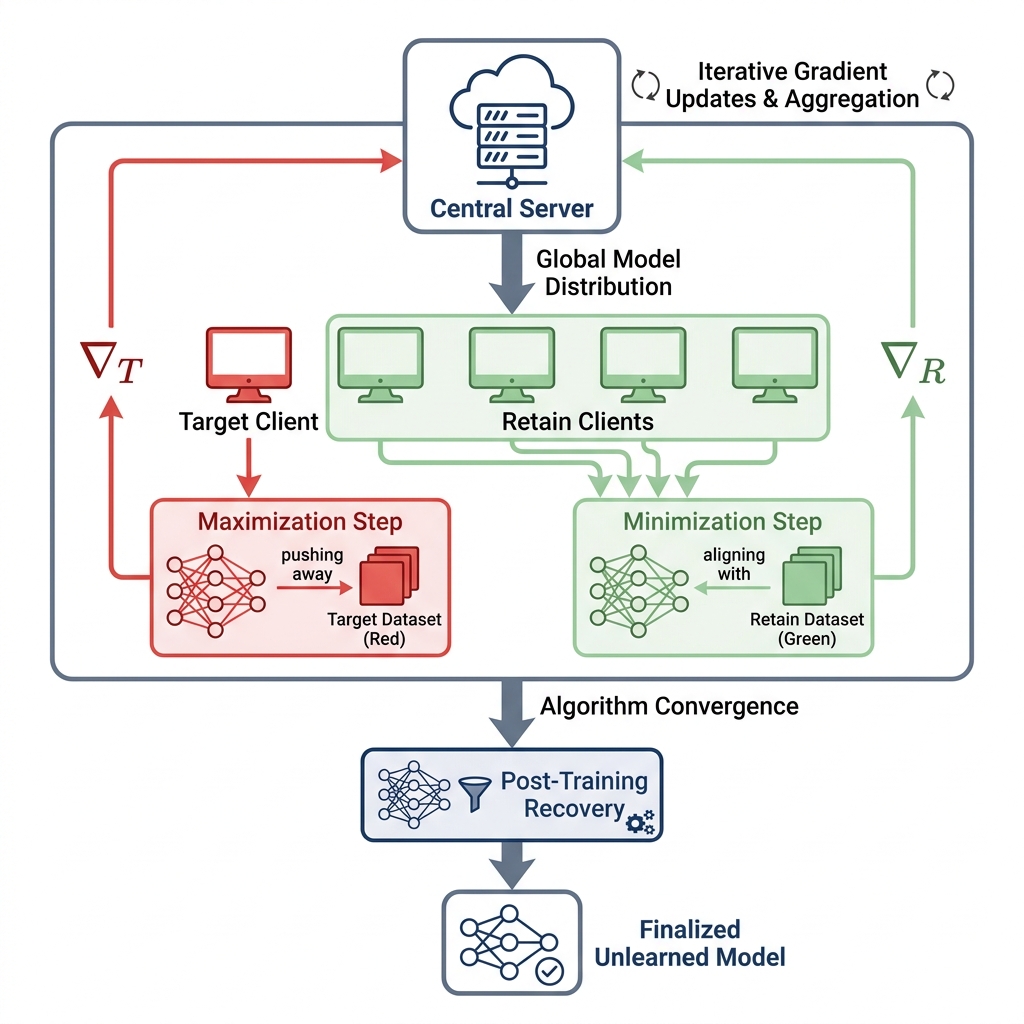}
        \caption{Client-level unlearning.}
        \label{fig:client-level-unlearning}
    \end{subfigure}
    \hfill
    \begin{subfigure}[t]{0.47\textwidth}
        \centering
        \vspace{0pt}
        \includegraphics[
            height=0.245\textheight,
            keepaspectratio,
            trim=10 10 10 10,
            clip
        ]{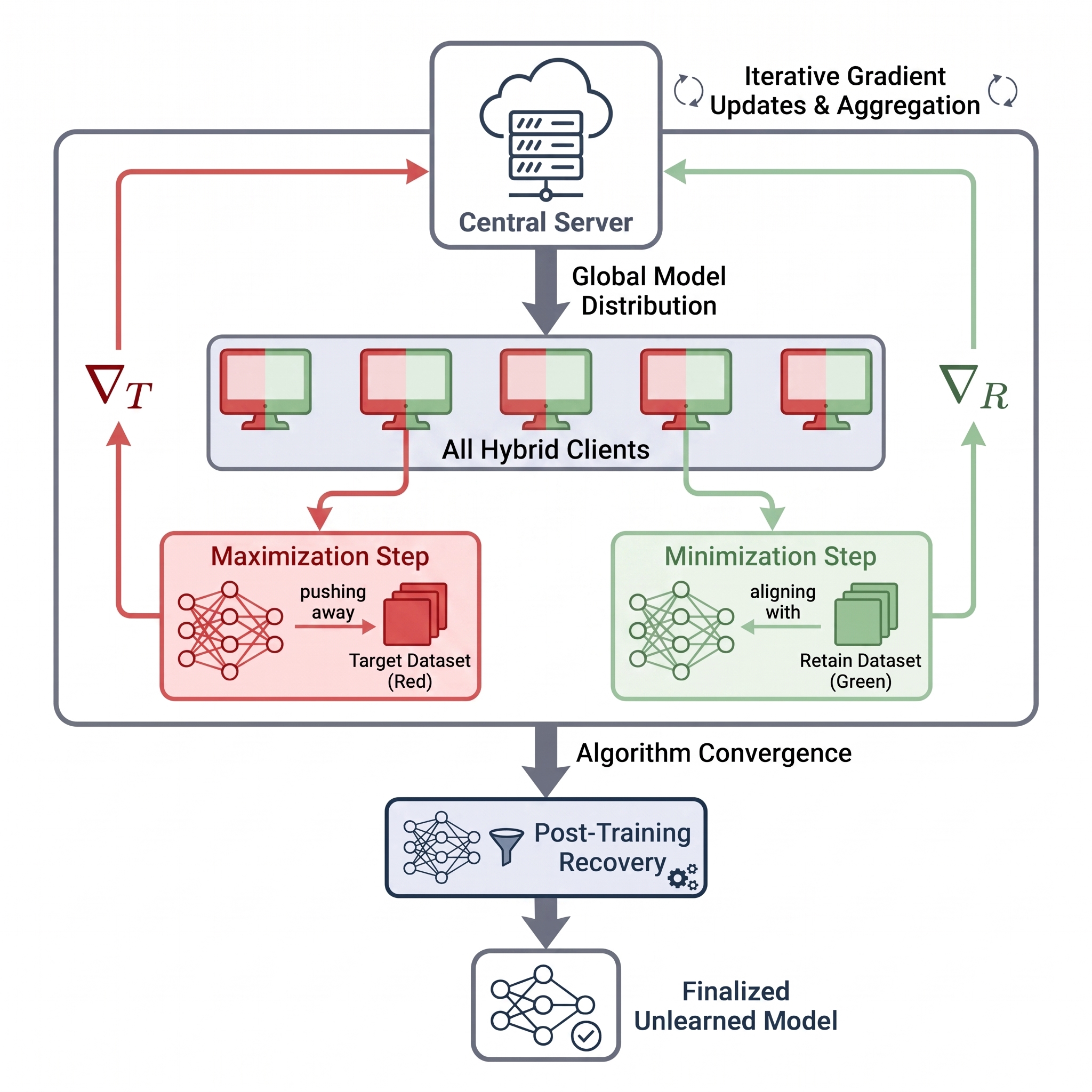}
        \caption{Data-level unlearning.}
        \label{fig:data-level-unlearning}
    \end{subfigure}

    \caption{
    Overview of \method{} under client-level and data-level unlearning.
    In client-level unlearning, the contribution of an entire target client is
    removed, whereas in data-level unlearning only selected samples are forgotten
    while the corresponding clients remain active. In both settings, forget-set
    maximization and retained-data minimization alternate for $T$ rounds,
    followed by retained-data post-training.
    }
    \label{fig:fed-f-scrub-overview}
\end{figure*}

 Let  $\theta$ denote the student model parameters. and $\theta_T$ be a reference model. Inspired by min-max problem, for client $c$, we define,
 \begingroup
\setlength{\abovedisplayskip}{4pt}
\setlength{\belowdisplayskip}{4pt}
\setlength{\jot}{2pt}
\begin{align*}
\mathcal{L}_{\max}^{(c)}(S_F^{(c)},\theta,\theta_T)
&=\frac{1}{n_F^{(c)}}\sum_{x_f\in S_F^{(c)}} d_{f_F}(x_f;\theta,\theta_T),\\
\mathcal{L}_{\min}^{(c)}(S_R^{(c)},\theta,\theta_T)
&=\frac{\alpha}{n_R^{(c)}}\sum_{x_r\in S_R^{(c)}} d_{f_R}(x_r;\theta,\theta_T)
+\frac{\gamma}{n_R^{(c)}}\sum_{(x_r,y_r)\in S_R^{(c)}} \ell_{f_R}(h(x_r;\theta),y_r),
\end{align*}
\endgroup
 where $\alpha$ and $\gamma$ are loss weights. 
Inspired by a min-max objective, we aim to maximize the $f$-divergence between the outputs of the unlearned model and the reference model on the forgetting dataset, while simultaneously minimizing the $f$-divergence between their outputs on the remaining dataset. The  \method  algorithm is presented in Algorithm~\ref{alg:fed-f-scrub-gradient-local} (provided in Appendix ~\ref{appendix:Alg}). We further studied \method from \textit{theoretical perspective} in Appendix ~\ref{sec:theory-res}, where we study the choice of $f$-divergence for forgetting, and derive a local curvature characterization that makes the role of the chosen divergence explicit. Next, we provide a toy example for motivation of our min-max algorithm,  \method algorithm.

\textbf{Toy Example; Min-Max Unlearning Dynamics:} We use a 2D toy classification task to visualize why  \method requires both
alternating min-max optimization and post-training recovery. The retained set
$S_R$ consists of two well-separated classes, while the forget set $S_F$
contains maliciously labeled borderline samples placed near the decision
boundary. Pre-training on this poisoned data warps the boundary toward the
wrong class.

\begin{figure*}[t]
    \centering
    \includegraphics[width=0.9\textwidth]{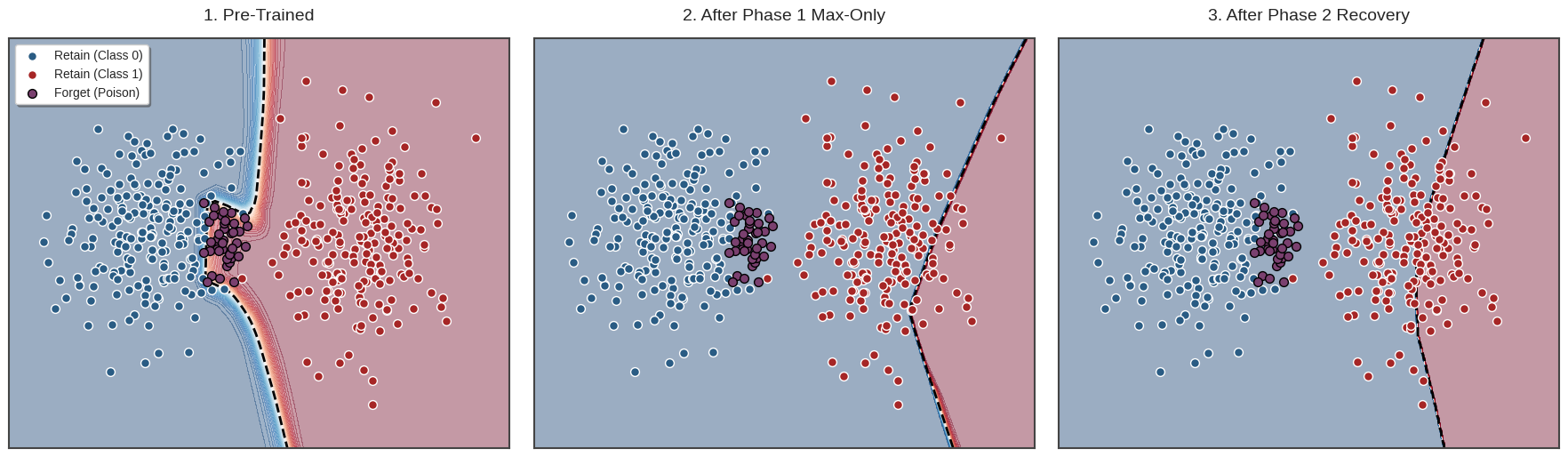}
    \caption{
        Catastrophic collapse under unconstrained forgetting.
        \textbf{Left:} the pre-trained model memorizes the poisoned borderline
        samples. \textbf{Middle:} f-divergence maximization without a retention
        anchor collapses the boundary and misclassifies clean samples.
        \textbf{Right:} recovery must effectively retrain from the degraded state.
    }
    \label{fig:toy_max_only_ablation}
\end{figure*}

Figure~\ref{fig:toy_max_only_ablation} shows that using only the maximization step leads to catastrophic collapse. Although the
poisoned samples are pushed away from the learned hypothesis, the absence of a
retention constraint causes the decision boundary to move indiscriminately,
misclassifying clean retained samples. Subsequent recovery therefore resembles
retraining rather than efficient unlearning.

\begin{figure*}[t]
    \centering
    \includegraphics[width=0.9\textwidth]{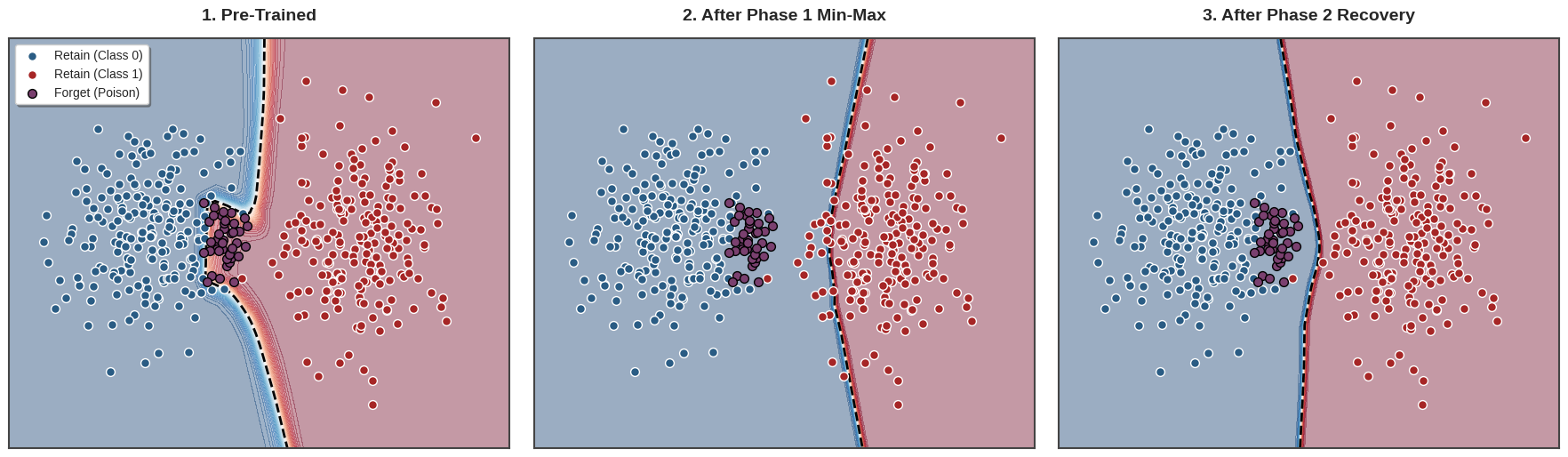}
    \caption{
        Stable unlearning under  \method.
        \textbf{Left:} the poisoned samples distort the pre-trained boundary.
        \textbf{Middle:} alternating min-max optimization removes the poison
        while retaining the global boundary structure. \textbf{Right:}
        post-training recovery smooths the residual distortion and restores
        clean class separation.
    }
    \label{fig:toy_boundary_dynamics}
\end{figure*}

In contrast, Figure~\ref{fig:toy_boundary_dynamics} shows that max-min avoids
this collapse. During Phase 1, the maximization step removes the influence of
$S_F$, while the minimization step on $S_R$ anchors the boundary and preserves
the global structure of the model. This produces only a localized temporary
distortion. During Phase 2, post-training minimization smooths this distortion
and restores a clean decision boundary. These results illustrate that the
alternating minimization step is essential for stable unlearning, while the
post-training phase acts as lightweight recovery rather than full retraining.

\section{Experiments}\label{sec:experiments}
\textbf{Datasets \& Models.}
We conduct experiments on CIFAR-10/100, FashionMNIST, and MNIST using three model architectures: (i) ResNet-18, (ii) a lightweight CNN consisting of two convolutional layers with Layer Normalization, and (iii) LeNet5. All experiments were conducted on a cluster with one NVIDIA H100 GPU, 10 CPU cores, and 512 GB RAM. Unless stated otherwise, each reported unlearning result is averaged over three independent runs of the unlearning procedure.

\paragraph{Experimental repetitions and model selection.}
Within each dataset and federation setting, all methods are assigned
the same predefined budget $B$ (Appendix \ref{sec:implementation-details}) of server-client synchronization
phases. For \method, a forget-retain iteration uses two phases and
each post-training round uses one, so $B=2\times T+T_{\mathrm{post}}$.
Among checkpoints produced within this budget, we select the one with
the highest clean accuracy on a held-out validation set. We evaluate
all reported metrics at that same checkpoint. Neither test-set
performance nor forgetting or MIA metrics are used for selection.

Across runs, stochastic operations during unlearning, such as minibatch or client sampling where applicable, use independent random seeds. Consequently, the reported standard deviations characterize variability conditional on the fixed pretrained checkpoint and deletion request. MIA samples are also selected independently for each run, so the corresponding standard deviations additionally include evaluation-sampling variability.

\textbf{Baselines.}
We compare our method against:  (1) \textbf{NoT} \citep{khalil2025notfederatedunlearningweight}; (2) \textbf{MoDE} \citep{10269017}; and (3) \textbf{Halimi} \citep{halimi2023federatedunlearningefficientlyerase}. Our codes are available  \href{https://anonymous.4open.science/r/Federated_unlearning_minmax_f/Readme.md}{\textcolor{magenta}{here}}.


\textbf{Privacy Evaluation. } For privacy evaluation, we consider the membership inference attack (MIA) score for our models. We have different kind of MIAs, but we focused on the simple one and Yeom \citep{yeom2018privacyriskmachinelearning} for CIFAR-100. The attacks evaluate whether the model retains information about the forget set by comparing its behavior on forgotten samples and non-member test samples.
If the model treats forgotten data similarly to data it never saw, the unlearning was successful and privacy is preserved, with the membership inference accuracy approaching 50\% (random guessing) being the ideal outcome; however, if the accuracy significantly deviates from 50\%, this indicates the model's responses reveal distinguishable patterns for forgotten data, suggesting privacy leakage and incomplete removal of the data's influence. Another important fact is that under our setup we remove the clean data, based on the literature of Differential Privacy (DP) \citep{el2022differential}, we know that working with noisy data can help us to increase our privacy budget which should be avoided in the FU experiments.

\textbf{Utility Evaluation.} For the backdoor and label-confusion scenarios, we select checkpoints
using clean accuracy on a held-out validation set that is disjoint from
the optimization data. We then report clean accuracy for the selected
checkpoint on the separate official test set; test-set results are not
used for checkpoint selection. Clean test accuracy measures predictive
utility, while attack success rate on triggered test examples, where
reported, measures removal of backdoor behavior. Within each
experimental setting, all methods receive the same predefined budget
of server-client synchronization phases. Equal phase budgets do not
imply equal transmitted bits, local computation, or wall-clock time;
these quantities were not measured consistently for every baseline.

\begin{table*}[t]
\caption{CIFAR-100 results for client-level and data-level unlearning under backdoor, confused-data, and privacy scenarios.
For client-level unlearning, the setting denotes the number of clients $N$.
For data-level unlearning, $N/r$ denotes $N$ clients with $r\%$ of local data removed from each client.
Acc.: higher is better; ASR: lower is better; MIA is best when closest to 50.}
\label{tab:cifar100-combined}
\centering
\setlength{\tabcolsep}{3.0pt}
\renewcommand{\arraystretch}{1.08}
\resizebox{\textwidth}{!}{%
\begin{tabular}{l c l cccc cc ccccc}
\toprule
\multirow{2}{*}{\textbf{Scenario}}
& \multirow{2}{*}{\textbf{Setting}}
& \multirow{2}{*}{\textbf{Method}}
& \multicolumn{4}{c}{\textbf{Backdoor}}
& \multicolumn{2}{c}{\textbf{Confused}}
& \multicolumn{5}{c}{\textbf{Privacy}} \\
\cmidrule(lr){4-7}
\cmidrule(lr){8-9}
\cmidrule(lr){10-14}
& & &
\textbf{Pre. Acc.} & \textbf{Pre. ASR}
& \textbf{Unl. Acc.} & \textbf{Unl. ASR}
& \textbf{Pre. Acc.} & \textbf{Unl. Acc.}
& \textbf{Pre. Acc.} & \textbf{Unl. Acc.}
& \textbf{MIA} & \textbf{Yeom Pre.} & \textbf{Yeom Unl.} \\
\midrule


\multirow{8}{*}{\textbf{\shortstack{Client-\\Level}}}
& \multirow{4}{*}{5}
& Ours (KL)
& \multirow{4}{*}{47.90}
& \multirow{4}{*}{72.05}
& 50.10 $\pm$ 0.40
& 5.80 $\pm$ 0.54
& \multirow{4}{*}{45.62}
& 49.80 $\pm$ 0.44
& \multirow{4}{*}{53.08}
& \textbf{51.70 $\pm$ 0.53}
& 56.70 $\pm$ 0.56
& \multirow{4}{*}{65.75}
& \textbf{57.40 $\pm$ 1.10} \\
& & Ours (JS)
& &
& \textbf{50.22 $\pm$ 0.43}
& 7.98 $\pm$ 0.40
&
& 49.48 $\pm$ 0.32
&
& 51.44 $\pm$ 0.33
& \textbf{52.23 $\pm$ 0.56}
&
& 58.59 $\pm$ 0.17 \\
& & NoT
& &
& 44.40 $\pm$ 1.39
& \textbf{0.59 $\pm$ 0.14}
&
& 35.60 $\pm$ 1.25
&
& 37.74 $\pm$ 1.99
& 53.35 $\pm$ 1.03
&
& 59.32 $\pm$ 0.95 \\
& & Halimi
& &
& 48.01 $\pm$ 0.48
& 5.34 $\pm$ 0.90
&
& \textbf{49.92 $\pm$ 0.68}
&
& 51.00 $\pm$ 0.44
& 57.29 $\pm$ 0.62
&
& 60.73 $\pm$ 0.32 \\
\cmidrule(lr){2-14}


& \multirow{4}{*}{10}
& Ours (KL)
& \multirow{4}{*}{48.21}
& \multirow{4}{*}{77.24}
& \textbf{49.44 $\pm$ 0.72}
& 2.16 $\pm$ 0.69
& \multirow{4}{*}{45.32}
& \textbf{49.12 $\pm$ 0.48}
& \multirow{4}{*}{52.11}
& \textbf{51.38 $\pm$ 0.95}
& 58.00 $\pm$ 0.51
& \multirow{4}{*}{63.96}
& 57.92 $\pm$ 1.18 \\
& & Ours (JS)
& &
& 49.06 $\pm$ 0.11
& 1.04 $\pm$ 0.50
&
& 48.48 $\pm$ 0.13
&
& 51.36 $\pm$ 0.15
& 56.60 $\pm$ 0.25
&
& 57.71 $\pm$ 0.14 \\
& & NoT
& &
& 46.18 $\pm$ 2.14
& \textbf{0.38 $\pm$ 0.11}
&
& 42.28 $\pm$ 0.76
&
& 45.50 $\pm$ 0.94
& \textbf{49.20 $\pm$ 1.22}
&
& \textbf{52.52 $\pm$ 1.18} \\
& & Halimi
& &
& 49.40 $\pm$ 0.34
& 1.36 $\pm$ 0.64
&
& 48.42 $\pm$ 0.42
&
& 50.48 $\pm$ 1.05
& 61.00 $\pm$ 0.72
&
& 59.66 $\pm$ 0.58 \\
\midrule


\multirow{16}{*}{\textbf{\shortstack{Data-\\Level}}}
& \multirow{4}{*}{5/6}
& Ours (KL)
& \multirow{4}{*}{47.20}
& \multirow{4}{*}{80.61}
& \textbf{52.75 $\pm$ 0.68}
& 0.69 $\pm$ 0.34
& \multirow{4}{*}{51.24}
& \textbf{53.18 $\pm$ 1.13}
& \multirow{4}{*}{52.32}
& 51.14 $\pm$ 1.07
& \textbf{49.80 $\pm$ 0.59}
& \multirow{4}{*}{58.11}
& \textbf{51.17 $\pm$ 0.56} \\
& & Ours (JS)
& &
& 51.53 $\pm$ 0.28
& 0.77 $\pm$ 0.29
&
& 52.48 $\pm$ 0.18
&
& 50.98 $\pm$ 0.25
& 48.80 $\pm$ 0.25
&
& 51.36 $\pm$ 0.29 \\
& & NoT
& &
& 35.30 $\pm$ 1.08
& 2.63 $\pm$ 1.39
&
& 40.12 $\pm$ 1.43
&
& 49.24 $\pm$ 1.34
& 50.44 $\pm$ 0.95
&
& 52.18 $\pm$ 1.44 \\
& & Halimi
& &
& 52.10 $\pm$ 0.45
& \textbf{0.36 $\pm$ 0.08}
&
& 53.16 $\pm$ 0.51
&
& \textbf{52.56 $\pm$ 0.73}
& 54.44 $\pm$ 0.61
&
& 56.16 $\pm$ 0.38 \\
\cmidrule(lr){2-14}


& \multirow{4}{*}{5/10}
& Ours (KL)
& \multirow{4}{*}{45.16}
& \multirow{4}{*}{81.07}
& \textbf{51.03 $\pm$ 0.82}
& \textbf{0.34 $\pm$ 0.17}
& \multirow{4}{*}{49.42}
& \textbf{50.76 $\pm$ 0.96}
& \multirow{4}{*}{53.11}
& \textbf{51.05 $\pm$ 0.81}
& \textbf{49.89 $\pm$ 1.02}
& \multirow{4}{*}{59.08}
& \textbf{50.10 $\pm$ 0.58} \\
& & Ours (JS)
& &
& 50.35 $\pm$ 0.34
& 0.68 $\pm$ 0.31
&
& 49.65 $\pm$ 0.48
&
& 50.98 $\pm$ 0.24
& 52.60 $\pm$ 0.49
&
& 53.57 $\pm$ 0.43 \\
& & NoT
& &
& 38.58 $\pm$ 1.57
& 34.30 $\pm$ 1.05
&
& 39.93 $\pm$ 0.57
&
& 50.30 $\pm$ 0.84
& \textbf{50.11 $\pm$ 1.24}
&
& 49.63 $\pm$ 1.33 \\
& & Halimi
& &
& 50.23 $\pm$ 0.46
& 0.93 $\pm$ 0.41
&
& 50.00 $\pm$ 0.66
&
& 50.34 $\pm$ 0.70
& 56.33 $\pm$ 0.51
&
& 57.50 $\pm$ 0.70 \\
\cmidrule(lr){2-14}


& \multirow{4}{*}{10/6}
& Ours (KL)
& \multirow{4}{*}{47.62}
& \multirow{4}{*}{74.09}
& 48.00 $\pm$ 0.45
& 0.25 $\pm$ 0.05
& \multirow{4}{*}{48.43}
& \textbf{50.64 $\pm$ 1.29}
& \multirow{4}{*}{51.08}
& 48.64 $\pm$ 0.70
& \textbf{49.00 $\pm$ 0.95}
& \multirow{4}{*}{56.00}
& 51.46 $\pm$ 0.30 \\
& & Ours (JS)
& &
& 47.50 $\pm$ 0.04
& 0.25 $\pm$ 0.19
&
& 48.22 $\pm$ 0.23
&
& 47.98 $\pm$ 0.24
& 51.45 $\pm$ 0.20
&
& 51.86 $\pm$ 0.17 \\
& & NoT
& &
& 34.10 $\pm$ 0.93
& 21.06 $\pm$ 1.48
&
& 36.78 $\pm$ 1.06
&
& 46.30 $\pm$ 0.72
& 48.60 $\pm$ 1.96
&
& \textbf{49.94 $\pm$ 1.07} \\
& & Halimi
& &
& \textbf{50.28 $\pm$ 0.42}
& \textbf{0.20 $\pm$ 0.04}
&
& 49.60 $\pm$ 0.83
&
& \textbf{49.96 $\pm$ 0.91}
& 54.60 $\pm$ 0.37
&
& 54.25 $\pm$ 0.86 \\
\cmidrule(lr){2-14}


& \multirow{4}{*}{10/10}
& Ours (KL)
& \multirow{4}{*}{45.30}
& \multirow{4}{*}{80.61}
& \textbf{47.94 $\pm$ 1.46}
& \textbf{0.28 $\pm$ 0.08}
& \multirow{4}{*}{45.68}
& \textbf{48.20 $\pm$ 0.83}
& \multirow{4}{*}{50.48}
& \textbf{47.92 $\pm$ 0.64}
& 51.58 $\pm$ 1.20
& \multirow{4}{*}{55.47}
& 49.56 $\pm$ 0.79 \\
& & Ours (JS)
& &
& 45.88 $\pm$ 0.34
& 0.51 $\pm$ 0.32
&
& 45.70 $\pm$ 0.24
&
& 44.32 $\pm$ 0.39
& 48.83 $\pm$ 0.34
&
& \textbf{50.03 $\pm$ 0.24} \\
& & NoT
& &
& 35.24 $\pm$ 0.95
& 10.61 $\pm$ 0.84
&
& 37.83 $\pm$ 1.84
&
& 45.56 $\pm$ 1.39
& \textbf{50.00 $\pm$ 1.41}
&
& 49.84 $\pm$ 1.57 \\
& & Halimi
& &
& 46.96 $\pm$ 0.79
& 0.48 $\pm$ 0.13
&
& 47.18 $\pm$ 0.77
&
& 47.81 $\pm$ 0.98
& 53.10 $\pm$ 0.78
&
& 53.64 $\pm$ 0.81 \\
\bottomrule
\end{tabular}%
}
\end{table*}

\textbf{Client-Level Unlearning:} Following \citep{halimi2023federatedunlearningefficientlyerase}, we simulate full removal of one client, where the forgotten data may be \textbf{backdoored}, \textbf{label-confused}, or \textbf{clean} (in the privacy setting). Specifically, we consider two configurations: $N=5$ clients with one client holding $66\%$ corrupted data, and $N=10$ clients with one client holding $80\%$ corrupted data. 
Backdoor corruption follows the $3 \times 3$ pixel trigger pattern introduced by \citep{gu2019badnetsidentifyingvulnerabilitiesmachine}, while label confusion is implemented via deterministic class mapping to simulate annotation errors. The results are available in Table \ref{tab:cifar100-combined} and \ref{tab:client-level-backdoor-confused} . For each method, all reported metrics are computed at the checkpoint
selected by held-out validation accuracy from those produced within
its predefined synchronization-phase budget.
\begin{table*}[t]
\caption{Comparison of client-level unlearning performance under backdoor and confused-data scenarios. In each setting, one client is removed, with 66\% corrupted data when \(N=5\) and 80\% corrupted data when \(N=10\).}
\label{tab:client-level-backdoor-confused}
\centering
\setlength{\tabcolsep}{3.0pt}
\renewcommand{\arraystretch}{1.08}
\resizebox{\textwidth}{!}{%
\begin{tabular}{l c cc cc cc cc cc cc}
\toprule

\multirow{3}{*}{\textbf{Method}}
& \multirow{3}{*}{$N$}
& \multicolumn{6}{c}{\textbf{Backdoor}}
& \multicolumn{6}{c}{\textbf{Confused}} \\

\cmidrule(lr){3-8}
\cmidrule(lr){9-14}

&
& \multicolumn{2}{c}{\textbf{CIFAR-10}}
& \multicolumn{2}{c}{\textbf{MNIST}}
& \multicolumn{2}{c}{\textbf{Fashion-MNIST}}
& \multicolumn{2}{c}{\textbf{CIFAR-10}}
& \multicolumn{2}{c}{\textbf{MNIST}}
& \multicolumn{2}{c}{\textbf{Fashion-MNIST}} \\

\cmidrule(lr){3-4}
\cmidrule(lr){5-6}
\cmidrule(lr){7-8}
\cmidrule(lr){9-10}
\cmidrule(lr){11-12}
\cmidrule(lr){13-14}

&
& \textbf{Pre.} & \textbf{Unl.}
& \textbf{Pre.} & \textbf{Unl.}
& \textbf{Pre.} & \textbf{Unl.}
& \textbf{Pre.} & \textbf{Unl.}
& \textbf{Pre.} & \textbf{Unl.}
& \textbf{Pre.} & \textbf{Unl.} \\

\midrule


Ours (KL)
& \multirow{6}{*}{10}
& \multirow{6}{*}{63.64} & 71.84 $\pm$ 1.29
& \multirow{6}{*}{99.14} & 99.16 $\pm$ 0.34
& \multirow{6}{*}{73.85} & 73.10 $\pm$ 2.10
& \multirow{6}{*}{63.64} & 71.24 $\pm$ 3.20 
& \multirow{6}{*}{98.93} & 99.11 $\pm$ 0.12
& \multirow{6}{*}{78.28} & 77.62 $\pm$ 1.25 \\

Ours (JS)
&
& & 72.74 $\pm$ 0.19
& & 98.99 $\pm$ 0.27
& & \textbf{75.13} $\pm$ 1.45
& & 72.90 $\pm$ 0.44
& & 99.08 $\pm$ 0.19
& & \textbf{77.87} $\pm$1.02 \\

Ours ($\chi^2$)
&
& & \textbf{72.92} $\pm$ 0.89
& & \textbf{99.21} $\pm$ 0.29
& & 73.82  $\pm$ 1.04
& & \textbf{77.43} $\pm$ 1.23
& & 99.06 $\pm$ 0.27
& & 77.43 $\pm$ 1.51 \\

NoT
&
& & 67.00 $\pm$ 2.65
& & 82.60 $\pm$ 1.46
& & 46.82 $\pm$ 3.36
& & 69.20 $\pm$ 1.98
& & 88.31 $\pm$ 1.84
& & 76.07 $\pm$ 1.64 \\

Halimi
&
& & 72.56 $\pm$ 1.10
& & \textbf{99.21} $\pm$ 0.25
& & 73.32 $\pm$ 1.77
& & 72.66 $\pm$ 1.56
& & \textbf{99.13} $\pm$ 0.15
& & 77.73 $\pm$ 1.85 \\

MoDE
&
& & 70.96 $\pm$ 1.43
& & 98.74 $\pm$ 0.72
& & 72.07 $\pm$ 1.73
& & 70.96 $\pm$ 3.43
& & 81.02 $\pm$ 2.23
& & 77.68 $\pm$ 1.39 \\

\midrule


Ours (KL)
& \multirow{6}{*}{5}
& \multirow{6}{*}{52.22} & 69.62 $\pm$ 1.35
& \multirow{6}{*}{98.98} & 98.88 $\pm$ 0.54
& \multirow{6}{*}{31.32} & 39.55 $\pm$ 2.95
& \multirow{6}{*}{52.22} & 67.36 $\pm$ 1.17
& \multirow{6}{*}{97.93} & 98.66 $\pm$ 0.14
& \multirow{6}{*}{66.58} & 58.58 $\pm$  3.63\\

Ours (JS)
&
& & \textbf{77.50} $\pm$ 0.32
& & 98.81 $\pm$  0.09
& & 55.93 $\pm$ 1.08
& & \textbf{77.22} $\pm$ 0.91
& & 98.14 $\pm$  0.06
& & \textbf{78.12} $\pm$ 1.40 \\

Ours ($\chi^2$)
&
& & 77.22 $\pm$ 1.43
& & \textbf{98.95} $\pm$ 0.91
& & 48.48 $\pm$ 3.12
& & 61.07 $\pm$ 2.40
& & \textbf{99.13} $\pm$ 0.34
& & 61.07 $\pm$ 3.93 \\

NoT
&
& & 70.16 $\pm$ 1.98
& & 89.21 $\pm$ 2.66
& & 37.45 $\pm$ 2.51
& & 68.26 $\pm$ 1.35
& & 83.98 $\pm$ 2.61
& & 32.76 $\pm$ 4.54\\

Halimi
&
& & 66.76 $\pm$ 1.03
& & 98.81 $\pm$ 0.30
& & 54.02 $\pm$ 2.78
& & 66.76 $\pm$ 1.65
& & 98.93 $\pm$ 0.64
& & 62.60 $\pm$ 1.72\\

MoDE
&
& & 65.44 $\pm$ 1.78
& & 98.49 $\pm$ 0.63
& & \textbf{64.30} $\pm$ 3.57
& & 69.44 $\pm$ 1.82
& & 98.36 $\pm$ 0.84
& & 70.20 $\pm$ 3.33\\

\bottomrule
\end{tabular}%
}
\end{table*}

\textbf{Data-Level Unlearning:} Inspired by \citep{wang2024goldfishefficientfederatedunlearning}, we evaluate unlearning when each client requests partial deletion of their local data. In this setup, forgotten samples are either clean, backdoored, or label-confused, and we vary the number of clients ($N \in \{5, 10\}$) as well as the fraction of forgotten data per client (2\%). More fractions of forgotten data per client are provided in Appendix~\ref{App:E}. For CIFAR-100 removed percentage is 6\% and 10\% (Table \ref{tab:cifar100-combined}) This setting reflects more realistic distributed deletion requests involving small, scattered portions of client datasets. For privacy oriented case we do not corrupt data as we explained why in Client-Level Unlearning discussion. The results are available in Tables \ref{tab:cifar100-combined} and \ref{tab:data-level-backdoor-confused}. For the more difficult situations and unlearning settings, you can check their result in the Appendix \ref{App:E}.

\begin{table*}[t]
\caption{Comparison of data-level unlearning performance under backdoor and confused-data scenarios. In all settings, 2\% of the data is removed from each client.}
\label{tab:data-level-backdoor-confused}
\centering
\setlength{\tabcolsep}{3.2pt}
\renewcommand{\arraystretch}{1.08}
\resizebox{\textwidth}{!}{%
\begin{tabular}{l c cc cc cc cc cc}
\toprule

\multirow{3}{*}{\textbf{Method}}
& \multirow{3}{*}{$N$}
& \multicolumn{4}{c}{\textbf{Backdoor}}
& \multicolumn{6}{c}{\textbf{Confused}} \\

\cmidrule(lr){3-6}
\cmidrule(lr){7-12}

&
& \multicolumn{2}{c}{\textbf{CIFAR-10}}
& \multicolumn{2}{c}{\textbf{Fashion-MNIST}}
& \multicolumn{2}{c}{\textbf{CIFAR-10}}
& \multicolumn{2}{c}{\textbf{MNIST}}
& \multicolumn{2}{c}{\textbf{Fashion-MNIST}} \\

\cmidrule(lr){3-4}
\cmidrule(lr){5-6}
\cmidrule(lr){7-8}
\cmidrule(lr){9-10}
\cmidrule(lr){11-12}

&
& \textbf{Pre.} & \textbf{Unl.}
& \textbf{Pre.} & \textbf{Unl.}
& \textbf{Pre.} & \textbf{Unl.}
& \textbf{Pre.} & \textbf{Unl.}
& \textbf{Pre.} & \textbf{Unl.} \\

\midrule


Ours (KL)
& \multirow{5}{*}{10}
& \multirow{5}{*}{53.38} & 68.62 $\pm$ 1.54
& \multirow{5}{*}{49.07} & 69.70 $\pm$ 0.06
& \multirow{5}{*}{67.02} & 70.66 $\pm$ 0.15
& \multirow{5}{*}{93.49} & 97.75 $\pm$ 0.61
& \multirow{5}{*}{42.35} & 67.67 $\pm$ 0.34 \\

Ours (JS)
&
& & \textbf{73.40} $\pm$ 0.86
& & 76.70 $\pm$ 0.83
& & \textbf{73.98} $\pm$ 0.11
& & \textbf{98.81} $\pm$ 0.23
& & \textbf{77.40} $\pm$ 0.04 \\

Ours ($\chi^2$)
&
& & 57.86 $\pm$ 2.80
& & 61.10 $\pm$ 1.37
& & 69.10 $\pm$ 1.21
& & 97.33 $\pm$ 0.90
& & 64.60 $\pm$ 2.21 \\

NoT
&
& & 62.00 $\pm$ 2.12
& & 12.37 $\pm$ 4.32
& & 69.32 $\pm$ 1.61
& & 86.58 $\pm$ 2.92
& & 39.92 $\pm$ 5.13 \\

Halimi
&
& & 62.00 $\pm$ 1.27
& & \textbf{79.53} $\pm$ 0.39
& & 70.00 $\pm$ 0.62
& & 97.80 $\pm$ 0.76
& & 68.52 $\pm$ 0.60 \\

\midrule


Ours (KL)
& \multirow{5}{*}{5}
& \multirow{5}{*}{58.54} & 66.64 $\pm$ 0.94
& \multirow{5}{*}{45.42} & 71.58 $\pm$ 0.16
& \multirow{5}{*}{68.70} & 73.00 $\pm$ 0.74
& \multirow{5}{*}{84.06} & 95.69 $\pm$ 0.51
& \multirow{5}{*}{50.77} & 70.70 $\pm$ 0.20 \\

Ours (JS)
&
& & \textbf{74.50} $\pm$ 0.72
& & \textbf{77.95} $\pm$ 0.43
& & \textbf{76.62} $\pm$ 0.55
& & \textbf{98.41} $\pm$ 0.14
& & 77.85 $\pm$ 1.05 \\

Ours ($\chi^2$)
&
& & 67.26 $\pm$ 1.80
& & 48.17 $\pm$ 2.94
& & 72.12 $\pm$ 2.11
& & 94.05 $\pm$ 1.07
& & 67.80 $\pm$ 3.80 \\

NoT
&
& & 69.92 $\pm$ 1.93
& & 52.82 $\pm$ 1.78
& & 72.00 $\pm$ 2.00
& & 81.64 $\pm$ 3.91
& & 68.40 $\pm$ 2.91 \\

Halimi
&
& & 68.00 $\pm$ 0.96
& & 71.22 $\pm$ 0.58
& & 74.00 $\pm$ 0.83
& & 73.52 $\pm$ 1.16
& & \textbf{95.48} $\pm$ 0.62 \\

\bottomrule
\end{tabular}%
}
\end{table*}

\begin{table*}[t]
\caption{Data-level privacy-oriented unlearning: pretrained (Pre.), unlearned (Unl.) accuracy, and MIA performance when 2\% of clean data is removed from each client. MoDE is omitted because it is explicitly designed for client-level unlearning.}
\label{tab:privacy-unlearning-1}
\centering
\setlength{\tabcolsep}{3.2pt}
\renewcommand{\arraystretch}{1.08}
\resizebox{\textwidth}{!}{%
\begin{tabular}{l c ccc ccc ccc}
\toprule
\multirow{2}{*}{\textbf{Method}}
& \multirow{2}{*}{$N$}
& \multicolumn{3}{c}{\textbf{CIFAR-10}}
& \multicolumn{3}{c}{\textbf{MNIST}}
& \multicolumn{3}{c}{\textbf{Fashion-MNIST}} \\
\cmidrule(lr){3-5}
\cmidrule(lr){6-8}
\cmidrule(lr){9-11}
&
& \textbf{Pre.} & \textbf{Unl.} & \textbf{MIA}
& \textbf{Pre.} & \textbf{Unl.} & \textbf{MIA}
& \textbf{Pre.} & \textbf{Unl.} & \textbf{MIA} \\
\midrule

Ours (KL)
& \multirow{5}{*}{10}
& \multirow{5}{*}{79.00} & 78.38 $\pm$ 0.66 & 60.77 $\pm$ 0.90
& \multirow{5}{*}{99.18} & 77.40 $\pm$ 1.25 & 59.89 $\pm$ 6.50
& \multirow{5}{*}{80.53} & 79.62 $\pm$ 0.11 & 49.90 $\pm$ 0.26 \\

Ours (JS)
& & & 78.32 $\pm$ 0.34 & 58.44 $\pm$ 1.10
& & \textbf{77.86} $\pm$ 0.87 & 59.33 $\pm$ 4.32
& & \textbf{80.02} $\pm$ 0.34 & \textbf{50.20} $\pm$ 0.12 \\

Ours ($\chi^2$)
& & & 78.12 $\pm$ 1.83 & 60.56 $\pm$ 1.43
& & 77.56 $\pm$ 1.63 & 59.78 $\pm$ 6.78
& & 79.28 $\pm$ 0.87 & 49.70 $\pm$ 0.55 \\

NoT
& & & 73.30 $\pm$ 2.31 & \textbf{49.66} $\pm$ 3.45
& & 72.92 $\pm$ 2.73 & \textbf{50.11} $\pm$ 2.33
& & 57.77 $\pm$ 4.50 & 49.50 $\pm$ 2.73 \\

Halimi
& & & \textbf{79.02} $\pm$ 0.86 & 59.33 $\pm$ 2.34
& & 77.72 $\pm$ 1.03 & 59.67 $\pm$ 3.02
& & 79.65 $\pm$ 0.27 & 49.20 $\pm$ 0.47 \\

\midrule

Ours (KL)
& \multirow{5}{*}{5}
& \multirow{5}{*}{80.00} & \textbf{78.68} $\pm$ 0.81 & 58.55 $\pm$ 0.56
& \multirow{5}{*}{98.93} & 77.66 $\pm$ 1.24 & 59.33 $\pm$ 5.55
& \multirow{5}{*}{79.92} & 79.33 $\pm$ 0.50 & 49.33 $\pm$ 0.19 \\

Ours (JS)
& & & 77.50 $\pm$ 0.23 & 59.00 $\pm$ 0.61
& & 77.72 $\pm$ 1.01 & 58.00 $\pm$ 4.37
& & \textbf{79.80} $\pm$ 0.19 & 48.38 $\pm$ 0.22 \\

Ours ($\chi^2$)
& & & 77.66 $\pm$ 0.90 & 60.44 $\pm$ 1.13
& & 77.52 $\pm$ 2.32 & 61.00 $\pm$ 5.18
& & 79.10 $\pm$ 1.15 & 48.57 $\pm$ 1.23 \\

NoT
& & & 72.78 $\pm$ 3.12 & \textbf{50.00} $\pm$ 0.00
& & 72.96 $\pm$ 2.45 & \textbf{50.89} $\pm$ 2.50
& & 72.03 $\pm$ 2.74 & 49.17 $\pm$ 2.76 \\

Halimi
& & & 77.98 $\pm$ 1.49 & 61.88 $\pm$ 2.23
& & \textbf{77.88} $\pm$ 1.17 & 61.89 $\pm$ 3.96
& & 79.18 $\pm$ 0.64 & \textbf{50.19} $\pm$ 0.75 \\

\bottomrule
\end{tabular}%
}
\end{table*}

\begin{table*}[t]
\caption{Client-level privacy-oriented unlearning: pretrained (Pre.), unlearned (Unl.) accuracy, and MIA performance when one client containing only clean data is removed.}
\label{tab:privacy-unlearning-2}
\centering
\setlength{\tabcolsep}{3.2pt}
\renewcommand{\arraystretch}{1.08}
\resizebox{\textwidth}{!}{%
\begin{tabular}{l c ccc ccc ccc}
\toprule
\multirow{2}{*}{\textbf{Method}}
& \multirow{2}{*}{$N$}
& \multicolumn{3}{c}{\textbf{CIFAR-10}}
& \multicolumn{3}{c}{\textbf{MNIST}}
& \multicolumn{3}{c}{\textbf{Fashion-MNIST}} \\
\cmidrule(lr){3-5}
\cmidrule(lr){6-8}
\cmidrule(lr){9-11}
&
& \textbf{Pre.} & \textbf{Unl.} & \textbf{MIA}
& \textbf{Pre.} & \textbf{Unl.} & \textbf{MIA}
& \textbf{Pre.} & \textbf{Unl.} & \textbf{MIA} \\
\midrule

Ours (KL)
& \multirow{6}{*}{10}
& \multirow{6}{*}{78.04} & 73.74 $\pm$ 2.40 & \textbf{52.40} $\pm$ 1.65
& \multirow{6}{*}{99.14} & 99.20 $\pm$ 0.03 & 50.33 $\pm$ 0.57
& \multirow{6}{*}{81.58} & \textbf{80.37} $\pm$ 2.15 & 52.67 $\pm$ 3.58 \\

Ours (JS)
& & & 63.16 $\pm$ 2.90 & 55.20 $\pm$ 5.10
& & 99.13 $\pm$ 0.04 & \textbf{50.00} $\pm$ 0.42
& & 80.25 $\pm$ 1.30 & 46.67 $\pm$ 2.35 \\

Ours ($\chi^2$)
& & & 77.66 $\pm$ 3.12 & 64.60 $\pm$ 3.40
& & \textbf{99.25} $\pm$ 0.06 & 50.50 $\pm$ 0.58
& & 79.67 $\pm$ 1.78 & 54.83 $\pm$ 2.50 \\

NoT
& & & 73.48 $\pm$ 1.83 & 52.80 $\pm$ 2.74
& & 95.25 $\pm$ 0.12 & 49.50 $\pm$ 0.85
& & 41.22 $\pm$ 0.95 & \textbf{48.67} $\pm$ 2.15 \\

Halimi
& & & \textbf{78.88} $\pm$ 1.96 & 64.80 $\pm$ 4.20
& & 99.21 $\pm$ 0.05 & \textbf{50.00} $\pm$ 1.50
& & 80.15 $\pm$ 1.12 & 54.00 $\pm$ 2.45 \\

MoDE
& & & 72.30 $\pm$ 3.12 & 64.20 $\pm$ 6.30
& & 98.98 $\pm$ 0.08 & 51.00 $\pm$ 2.72
& & 73.48 $\pm$ 1.60 & 44.83 $\pm$ 2.25 \\

\midrule

Ours (KL)
& \multirow{6}{*}{5}
& \multirow{6}{*}{79.72} & \textbf{78.92} $\pm$ 1.38 & 63.60 $\pm$ 6.25
& \multirow{6}{*}{98.98} & 98.88 $\pm$ 0.44 & 50.75 $\pm$ 0.12
& \multirow{6}{*}{81.33} & 80.80 $\pm$ 2.35 & 49.08 $\pm$ 1.05 \\

Ours (JS)
& & & 78.68 $\pm$ 0.87 & \textbf{49.90} $\pm$ 5.13
& & \textbf{99.13} $\pm$ 0.05 & \textbf{50.25} $\pm$ 0.30
& & \textbf{81.32} $\pm$ 1.25 & 48.42 $\pm$ 1.15 \\

Ours ($\chi^2$)
& & & 78.24 $\pm$ 2.10 & 63.40 $\pm$ 4.32
& & 98.88 $\pm$ 0.04 & 49.75 $\pm$ 0.26
& & 80.57 $\pm$ 1.85 & 50.75 $\pm$ 1.40 \\

NoT
& & & 72.94 $\pm$ 2.54 & 55.10 $\pm$ 3.45
& & 96.59 $\pm$ 1.15 & 48.75 $\pm$ 0.55
& & 38.95 $\pm$ 0.92 & 51.33 $\pm$ 1.70 \\

Halimi
& & & 77.52 $\pm$ 1.14 & 64.20 $\pm$ 4.56
& & 98.96 $\pm$ 0.06 & 50.67 $\pm$ 0.35
& & 80.65 $\pm$ 1.08 & \textbf{49.67} $\pm$ 1.25 \\

MoDE
& & & 70.60 $\pm$ 3.86 & 58.40 $\pm$ 4.62
& & 98.58 $\pm$ 0.49 & 49.12 $\pm$ 0.44
& & 68.77 $\pm$ 1.45 & 50.25 $\pm$ 1.85 \\

\bottomrule
\end{tabular}%
}
\end{table*}

\section{Discussion}

\textbf{The effect of choosing $f$-divergence:}
The choice of divergence changes the observed forgetting utility
trade-off, but no variant is best in every setting. In client-level
CIFAR-10 backdoor removal (Table~\ref{tab:client-level-backdoor-confused}),
KL-JS\footnote{In all configurations, the first loss component is fixed to KL.
Thus, table notation \textbf{Ours ($D$)} denotes $(\mathrm{KL},D,D)$;
e.g., \textbf{Ours (JS)} corresponds to KL-JS in the discussion.} achieves $72.74\%$ clean accuracy with $N=10$, compared with
$71.84\%$ for KL-KL and $72.92\%$ for KL-$\chi^2$. With $N=5$,
KL-JS achieves $77.50\%$, compared with $69.62\%$ for KL-KL and
$77.22\%$ for KL-$\chi^2$. CIFAR-100 shows why divergence choice
must also be evaluated against forgetting metrics
(Table~\ref{tab:cifar100-combined}). In client-level backdoor removal
with $N=10$, KL-JS achieves lower attack success rate (ASR) than
KL-KL ($1.04\%$ versus $2.16\%$) at similar clean accuracy. With \(N=5\), their ASR ordering reverses: KL-KL achieves 5.80\% ASR versus 7.98\% for KL-JS. These results support setting-dependent
divergence selection rather than a uniformly optimal choice. Although
JS divergence has a bounded value, that property alone does not
establish bounded parameter gradients or guarantee stable training.

\textbf{Data-level versus client-level unlearning:}
Client-level deletion removes one participant's contribution, whereas
data-level deletion targets a fraction of examples held by each
affected client. In the CIFAR-10 data-level backdoor setting, removing
$2\%$ of local data with $N=10$ raises clean accuracy from $53.38\%$
before unlearning to $73.40\%$ with KL-JS
(Table~\ref{tab:data-level-backdoor-confused}). Halimi reaches
$62.00\%$ in the same setting, so this result does not establish
dominance over the baselines. CIFAR-100 includes larger data-level
deletion fractions of $6\%$ and $10\%$
(Table~\ref{tab:cifar100-combined}). With five clients and $10\%$ of
local data deleted, KL-KL achieves $0.34\%$ ASR and $51.03\%$ clean
accuracy, compared with $34.30\%$ ASR and $38.58\%$ accuracy for NoT.
The CIFAR-100 results illustrate why we do not use ASR to
compare methods across datasets. In client-level unlearning with
$N=5$, NoT achieves a lower ASR than KL-KL (0.59\% versus 5.80\%),
but also lower clean test accuracy (44.40\% versus 50.10\%).
The same pattern appears with $N=10$: NoT achieves 0.38\% ASR
and 46.18\% accuracy, compared with 2.16\% ASR and 49.44\%
accuracy for KL-KL (Table~\ref{tab:cifar100-combined}). A low ASR can arise when a model
rarely predicts the attack target, even if its useful predictions
have deteriorated; these results do not establish that the model has unlearned the poison data. Motivated by this trade-off, we report
clean test accuracy for the other dataset comparisons and interpret
those results as comparisons of retained utility. 

For clean-data deletion, predictive utility and membership-inference
attack (MIA) accuracy should be considered together
(Tables~\ref{tab:cifar100-combined}, 
\ref{tab:privacy-unlearning-1}, \ref{tab:privacy-unlearning-2}). On CIFAR-100 client-level deletion with
$N=5$, KL-JS gives MIA accuracy of $52.23\%$, closer to random
guessing than KL-KL's $56.70\%$. With $N=10$, NoT's $49.20\%$ is
closer to random guessing than KL-JS's $56.60\%$. Thus, the privacy
results do not support a general advantage for one method across
deletion settings. MIA accuracy near $50\%$ measures the outcome of
the reported attack; it is not a guarantee that all information about
the deleted data has been removed.

\textbf{Comparison with baseline methods:}
The baselines reveal different trade-offs across datasets. For
client-level MNIST backdoor removal with $N=10$
(Table~\ref{tab:client-level-backdoor-confused}), NoT achieves
$82.60\%$ clean accuracy, while our KL-$\chi^2$ variant and Halimi
both achieve $99.21\%$. On CIFAR-100 client-level backdoor removal,
NoT achieves lower ASR than our variants but also lower clean
accuracy: for $N=5$ and $N=10$, its ASR is $0.59\%$ and $0.38\%$,
while its clean accuracy is $44.40\%$ and $46.18\%$, respectively
(Table~\ref{tab:cifar100-combined}). This pattern is consistent with
weight negation affecting useful predictions as well as backdoor
behavior, although these results do not isolate the cause of the
accuracy loss.

Halimi~\citep{halimi2023federatedunlearningefficientlyerase} is
competitive in several settings. On CIFAR-10 client-level backdoor
removal with $N=10$, it achieves $72.56\%$ clean accuracy, compared
with $72.74\%$ for our KL-JS variant. In contrast, for data-level
label-confusion removal on MNIST with $N=5$, KL-JS achieves $98.41\%$
versus Halimi's $73.52\%$
(Table~\ref{tab:data-level-backdoor-confused}). MoDE~\citep{10269017}
also has mixed results: on CIFAR-10 client-level backdoor removal
with $N=10$, it achieves $70.96\%$ versus $72.92\%$ for our
KL-$\chi^2$ variant, but on Fashion-MNIST with $N=5$ in client-level backdoor removal , MoDE achieves
$64.30\%$ versus $55.93\%$ for our best variant in that column
(Table~\ref{tab:client-level-backdoor-confused}). These comparisons
identify settings where \method performs well and settings where
a baseline performs better. Additional results appear in
Appendix~\ref{App:E}.

\section{Conclusion and Future Works}
We introduced \method, an active federated unlearning framework formulated as a min--max optimization problem using $f$-divergences such as Jensen--Shannon and $\chi^2$. The method increases the discrepancy between the unlearned and original models on forgotten data while preserving performance on retained data, thereby addressing both privacy-oriented and robustness-oriented unlearning in decentralized settings. Overall, \method provides a principled and flexible mechanism for balancing forgetting effectiveness and retained utility with limited trade-offs. Future work includes extending the framework to $\alpha$-R\'enyi divergences and combining it with the theoretical framework of \citep{ding2024understanding} to derive stronger guarantees.



\clearpage
\newpage

\subsection*{AI use statement}

Generative AI tools were used solely for limited editorial and presentation support. Specifically, they were used to assist with rewriting and improving the clarity, grammar, and presentation of portions of the manuscript, and to assist in generating and refining scientific figures based on descriptions provided by the authors.

All research ideas, methodological choices, mathematical derivations, theoretical results, experimental design, analyses, interpretations, and conclusions were developed and verified by the authors. The authors reviewed and take full responsibility for all AI-assisted text and figures included in the paper.

\subsection*{Ethics statement}

This work studies federated machine unlearning with the aim of reducing the influence of data requested for deletion while preserving model utility. Our experiments use only standard public benchmark datasets and simulated deletion, poisoning, and corruption scenarios; no private or personally identifiable user data are used. Membership-inference attacks are employed solely as evaluation tools and should not be interpreted as formal privacy guarantees. Likewise, the proposed method does not by itself certify exact deletion, differential privacy, or legal compliance with a right to erasure. All results should therefore be interpreted within the assumptions and evaluation settings stated in the paper.

\subsection*{Reproducibility statement}
All experiments use publicly available benchmark datasets, including CIFAR-10,
CIFAR-100, MNIST, and FashionMNIST. We report the model architectures,
federated-learning configurations, client- and data-level unlearning settings,
data-corruption procedures, evaluation metrics, and optimization
hyperparameters used in the experiments. Implementation details such as batch
size, learning rates, numbers of pre-training and unlearning rounds,
maximization/minimization schedules, and post-training configurations are
provided in the appendix. The implementation is based on the Flower federated-learning framework. Code
for reproducing the experiments is provided through the anonymous repository
accompanying the submission. The repository contains the implementation of the
proposed method and the experimental pipeline used to generate the reported
results.




\bibliography{Refs}

@article{polyanskiy2014lecture,
  title={Lecture notes on information theory},
  author={Polyanskiy, Yury and Wu, Yihong},
  journal={Lecture Notes for ECE563 (UIUC) and},
  volume={6},
  number={2012-2016},
  pages={7},
  year={2014},
  publisher={Citeseer}
}

@article{ding2024understanding,
  title={Understanding Fine-tuning in Approximate Unlearning: A Theoretical Perspective},
  author={Ding, Meng and Sharma, Rohan and Chen, Changyou and Xu, Jinhui and Ji, Kaiyi},
  journal={arXiv preprint arXiv:2410.03833},
  year={2024}
}

@misc{beutel2022flowerfriendlyfederatedlearning,
      title={Flower: A Friendly Federated Learning Research Framework}, 
      author={Daniel J. Beutel and Taner Topal and Akhil Mathur and Xinchi Qiu and Javier Fernandez-Marques and Yan Gao and Lorenzo Sani and Kwing Hei Li and Titouan Parcollet and Pedro Porto Buarque de Gusmão and Nicholas D. Lane},
      year={2022},
      eprint={2007.14390},
      archivePrefix={arXiv},
      primaryClass={cs.LG},
      url={https://arxiv.org/abs/2007.14390}, 
}

@misc{khalil2025notfederatedunlearningweight,
      title={NoT: Federated Unlearning via Weight Negation}, 
      author={Yasser H. Khalil and Leo Brunswic and Soufiane Lamghari and Xu Li and Mahdi Beitollahi and Xi Chen},
      year={2025},
      eprint={2503.05657},
      archivePrefix={arXiv},
      primaryClass={cs.LG},
      url={https://arxiv.org/abs/2503.05657}, 
}

@misc{hayes2024,
      title={Inexact Unlearning Needs More Careful Evaluations to Avoid a False Sense of Privacy}, 
      author={Jamie Hayes and Ilia Shumailov and Eleni Triantafillou and Amr Khalifa and Nicolas Papernot},
      year={2024},
      eprint={2403.01218},
      archivePrefix={arXiv},
      primaryClass={cs.LG},
      url={https://arxiv.org/abs/2403.01218}, 
}

@article{wang2024llm,
  title={LLM Unlearning via Loss Adjustment with Only Forget Data},
  author={Wang, Yaxuan and Wei, Jiaheng and Liu, Chris Yuhao and Pang, Jinlong and Liu, Quan and Shah, Ankit Parag and Bao, Yujia and Liu, Yang and Wei, Wei},
  journal={arXiv preprint arXiv:2410.11143},
  year={2024}
}

@inproceedings{bonato2025retain,
  title={Is Retain Set All You Need in Machine Unlearning? Restoring Performance of Unlearned Models with Out-Of-Distribution Images},
  author={Bonato, Jacopo and Cotogni, Marco and Sabetta, Luigi},
  booktitle={European Conference on Computer Vision},
  pages={1--19},
  year={2025},
  organization={Springer}
}

@article{jeon2024information,
  title={An Information Theoretic Metric for Evaluating Unlearning Models},
  author={Jeon, Dongjae and Jeung, Wonje and Kim, Taeheon and No, Albert and Choi, Jonghyun},
  journal={arXiv preprint arXiv:2405.17878},
  year={2024}
}

@article{rangel2024learning,
  title={Learning to Forget using Hypernetworks},
  author={Rangel, Jose Miguel Lara and Schoepf, Stefan and Foster, Jack and Krueger, David and Anwar, Usman},
  journal={arXiv preprint arXiv:2412.00761},
  year={2024}
}

@misc{georgiev2024attributetodeletemachineunlearningdatamodel,
      title={Attribute-to-Delete: Machine Unlearning via Datamodel Matching}, 
      author={Kristian Georgiev and Roy Rinberg and Sung Min Park and Shivam Garg and Andrew Ilyas and Aleksander Madry and Seth Neel},
      year={2024},
      eprint={2410.23232},
      archivePrefix={arXiv},
      primaryClass={cs.LG},
      url={https://arxiv.org/abs/2410.23232}, 
}

@misc{aminian2024robustsemisupervisedlearningfdivergence,
      title={Robust Semi-supervised Learning via $f$-Divergence and $\alpha$-R\'enyi Divergence}, 
      author={Gholamali Aminian and Amirhossien Bagheri and Mahyar JafariNodeh and Radmehr Karimian and Mohammad-Hossein Yassaee},
      year={2024},
      eprint={2405.00454},
      archivePrefix={arXiv},
      primaryClass={cs.LG},
      url={https://arxiv.org/abs/2405.00454}, 
}

@inproceedings{cao2015towards,
  title={Towards making systems forget with machine unlearning},
  author={Cao, Yinzhi and Yang, Junfeng},
  booktitle={2015 IEEE symposium on security and privacy},
  pages={463--480},
  year={2015},
  organization={IEEE}
}

@inproceedings{thudi2022unrolling,
  title={Unrolling sgd: Understanding factors influencing machine unlearning},
  author={Thudi, Anvith and Deza, Gabriel and Chandrasekaran, Varun and Papernot, Nicolas},
  booktitle={2022 IEEE 7th European Symposium on Security and Privacy (EuroS\&P)},
  pages={303--319},
  year={2022},
  organization={IEEE}
}

@article{nguyen2020variational,
  title={Variational bayesian unlearning},
  author={Nguyen, Quoc Phong and Low, Bryan Kian Hsiang and Jaillet, Patrick},
  journal={Advances in Neural Information Processing Systems},
  volume={33},
  pages={16025--16036},
  year={2020}
}

@article{guo2019certified,
  title={Certified data removal from machine learning models},
  author={Guo, Chuan and Goldstein, Tom and Hannun, Awni and Van Der Maaten, Laurens},
  journal={arXiv preprint arXiv:1911.03030},
  year={2019}
}

@misc{sekhari2021rememberwantforgetalgorithms,
      title={Remember What You Want to Forget: Algorithms for Machine Unlearning}, 
      author={Ayush Sekhari and Jayadev Acharya and Gautam Kamath and Ananda Theertha Suresh},
      year={2021},
      eprint={2103.03279},
      archivePrefix={arXiv},
      primaryClass={cs.LG},
      url={https://arxiv.org/abs/2103.03279}, 
}

@misc{bottou2018optimizationmethodslargescalemachine,
      title={Optimization Methods for Large-Scale Machine Learning}, 
      author={Léon Bottou and Frank E. Curtis and Jorge Nocedal},
      year={2018},
      eprint={1606.04838},
      archivePrefix={arXiv},
      primaryClass={stat.ML},
      url={https://arxiv.org/abs/1606.04838}, 
}

@misc{ghadimi2013stochasticfirstzerothordermethods,
      title={Stochastic First- and Zeroth-order Methods for Nonconvex Stochastic Programming}, 
      author={Saeed Ghadimi and Guanghui Lan},
      year={2013},
      eprint={1309.5549},
      archivePrefix={arXiv},
      primaryClass={math.OC},
      url={https://arxiv.org/abs/1309.5549}, 
}

@misc{hayes2024inexactunlearningneedscareful,
      title={Inexact Unlearning Needs More Careful Evaluations to Avoid a False Sense of Privacy}, 
      author={Jamie Hayes and Ilia Shumailov and Eleni Triantafillou and Amr Khalifa and Nicolas Papernot},
      year={2024},
      eprint={2403.01218},
      archivePrefix={arXiv},
      primaryClass={cs.LG},
      url={https://arxiv.org/abs/2403.01218}, 
}

@article{bartlett2006convexity,
  title={Convexity, Classification, and Risk Bounds},
  author={Bartlett, Peter L. and Jordan, Michael I. and McAuliffe, Jon D.},
  journal={Journal of the American Statistical Association},
  volume={101},
  number={473},
  pages={138--156},
  year={2006},
  publisher={Taylor \& Francis},
  doi={10.1198/016214505000000907}
}

@article{Nguyen_2010,
   title={Estimating Divergence Functionals and the Likelihood Ratio by Convex Risk Minimization},
   volume={56},
   ISSN={1557-9654},
   url={http://dx.doi.org/10.1109/TIT.2010.2068870},
   DOI={10.1109/tit.2010.2068870},
   number={11},
   journal={IEEE Transactions on Information Theory},
   publisher={Institute of Electrical and Electronics Engineers (IEEE)},
   author={Nguyen, XuanLong and Wainwright, Martin J. and Jordan, Michael I.},
   year={2010},
   month=Nov, pages={5847–5861} }

@article{li2024machine,
  title={Machine Unlearning: Taxonomy, Metrics, Applications, Challenges, and Prospects},
  author={Li, Na and Zhou, Chunyi and Gao, Yansong and Chen, Hui and Fu, Anmin and Zhang, Zhi and Shui, Yu},
  journal={arXiv preprint arXiv:2403.08254},
  year={2024}
}

@misc{kurmanji2023unboundedmachineunlearning,
      title={Towards Unbounded Machine Unlearning}, 
      author={Meghdad Kurmanji and Peter Triantafillou and Jamie Hayes and Eleni Triantafillou},
      year={2023},
      eprint={2302.09880},
      archivePrefix={arXiv},
      primaryClass={cs.LG},
      url={https://arxiv.org/abs/2302.09880}, 
}

@misc{roulet2025lossfunctionsoperatorsgenerated,
      title={Loss Functions and Operators Generated by f-Divergences}, 
      author={Vincent Roulet and Tianlin Liu and Nino Vieillard and Michael E. Sander and Mathieu Blondel},
      year={2025},
      eprint={2501.18537},
      archivePrefix={arXiv},
      primaryClass={cs.LG},
      url={https://arxiv.org/abs/2501.18537}, 
}

@article{hu2022membership,
  title={Membership inference attacks on machine learning: A survey},
  author={Hu, Hongsheng and Salcic, Zoran and Sun, Lichao and Dobbie, Gillian and Yu, Philip S and Zhang, Xuyun},
  journal={ACM Computing Surveys (CSUR)},
  volume={54},
  number={11s},
  pages={1--37},
  year={2022},
  publisher={ACM New York, NY}
}

@article{romandini2024federated,
  title={Federated unlearning: A survey on methods, design guidelines, and evaluation metrics},
  author={Romandini, Nicol{\`o} and Mora, Alessio and Mazzocca, Carlo and Montanari, Rebecca and Bellavista, Paolo},
  journal={IEEE Transactions on Neural Networks and Learning Systems},
  year={2024},
  publisher={IEEE}
}

@inproceedings{mcmahan2017communication,
  title={Communication-efficient learning of deep networks from decentralized data},
  author={McMahan, Brendan and Moore, Eider and Ramage, Daniel and Hampson, Seth and y Arcas, Blaise Aguera},
  booktitle={Artificial intelligence and statistics},
  pages={1273--1282},
  year={2017},
  organization={PMLR}
}

@misc{novello2024fdivergencebasedclassificationuse,
      title={$f$-Divergence Based Classification: Beyond the Use of Cross-Entropy}, 
      author={Nicola Novello and Andrea M. Tonello},
      year={2024},
      eprint={2401.01268},
      archivePrefix={arXiv},
      primaryClass={cs.LG},
      url={https://arxiv.org/abs/2401.01268}, 
}

@misc{goodfellow2014generativeadversarialnetworks,
      title={Generative Adversarial Networks}, 
      author={Ian J. Goodfellow and Jean Pouget-Abadie and Mehdi Mirza and Bing Xu and David Warde-Farley and Sherjil Ozair and Aaron Courville and Yoshua Bengio},
      year={2014},
      eprint={1406.2661},
      archivePrefix={arXiv},
      primaryClass={stat.ML},
      url={https://arxiv.org/abs/1406.2661}, 
}

@misc{nowozin2016fgantraininggenerativeneural,
      title={f-GAN: Training Generative Neural Samplers using Variational Divergence Minimization}, 
      author={Sebastian Nowozin and Botond Cseke and Ryota Tomioka},
      year={2016},
      eprint={1606.00709},
      archivePrefix={arXiv},
      primaryClass={stat.ML},
      url={https://arxiv.org/abs/1606.00709}, 
}

@misc{reid2009informationdivergenceriskbinary,
      title={Information, Divergence and Risk for Binary Experiments}, 
      author={Mark D. Reid and Robert C. Williamson},
      year={2009},
      eprint={0901.0356},
      archivePrefix={arXiv},
      primaryClass={stat.ML},
      url={https://arxiv.org/abs/0901.0356}, 
}

@misc{duchi2020learningmodelsuniformperformance,
      title={Learning Models with Uniform Performance via Distributionally Robust Optimization}, 
      author={John Duchi and Hongseok Namkoong},
      year={2020},
      eprint={1810.08750},
      archivePrefix={arXiv},
      primaryClass={stat.ML},
      url={https://arxiv.org/abs/1810.08750}, 
}

@misc{wang2024goldfishefficientfederatedunlearning,
      title={Goldfish: An Efficient Federated Unlearning Framework}, 
      author={Houzhe Wang and Xiaojie Zhu and Chi Chen and Paulo Esteves-Veríssimo},
      year={2024},
      eprint={2404.03180},
      archivePrefix={arXiv},
      primaryClass={cs.LG},
      url={https://arxiv.org/abs/2404.03180}, 
}

@misc{halimi2023federatedunlearningefficientlyerase,
      title={Federated Unlearning: How to Efficiently Erase a Client in FL?}, 
      author={Anisa Halimi and Swanand Kadhe and Ambrish Rawat and Nathalie Baracaldo},
      year={2023},
      eprint={2207.05521},
      archivePrefix={arXiv},
      primaryClass={cs.LG},
      url={https://arxiv.org/abs/2207.05521}, 
}

@misc{yeom2018privacyriskmachinelearning,
      title={Privacy Risk in Machine Learning: Analyzing the Connection to Overfitting}, 
      author={Samuel Yeom and Irene Giacomelli and Matt Fredrikson and Somesh Jha},
      year={2018},
      eprint={1709.01604},
      archivePrefix={arXiv},
      primaryClass={cs.CR},
      url={https://arxiv.org/abs/1709.01604}, 
}

@misc{dhasade2024quickdropefficientfederatedunlearning,
      title={QuickDrop: Efficient Federated Unlearning by Integrated Dataset Distillation}, 
      author={Akash Dhasade and Yaohong Ding and Song Guo and Anne-marie Kermarrec and Martijn De Vos and Leijie Wu},
      year={2024},
      eprint={2311.15603},
      archivePrefix={arXiv},
      primaryClass={cs.LG},
      url={https://arxiv.org/abs/2311.15603}, 
}

@INPROCEEDINGS{10682764,
  author={Meerza, Syed Irfan Ali and Sadovnik, Amir and Liu, Jian},
  booktitle={2024 IEEE Computer Society Annual Symposium on VLSI (ISVLSI)}, 
  title={ConFUSE: Confusion-based Federated Unlearning with Salience Exploration}, 
  year={2024},
  volume={},
  number={},
  pages={427-432},
  doi={10.1109/ISVLSI61997.2024.00083}}

@misc{jin2024forgettablefederatedlinearlearning,
      title={Forgettable Federated Linear Learning with Certified Data Unlearning}, 
      author={Ruinan Jin and Minghui Chen and Qiong Zhang and Xiaoxiao Li},
      year={2024},
      eprint={2306.02216},
      archivePrefix={arXiv},
      primaryClass={cs.LG},
      url={https://arxiv.org/abs/2306.02216}, 
}

@inproceedings{xiong2024appro,
  title={Appro-Fun: Approximate Machine Unlearning in Federated Setting},
  author={Xiong, Zuobin and Li, Wei and Cai, Zhipeng},
  booktitle={2024 33rd International Conference on Computer Communications and Networks (ICCCN)},
  pages={1--9},
  year={2024},
  organization={IEEE}
}

@inproceedings{xiong2023exact,
  title={Exact-fun: an exact and efficient federated unlearning approach},
  author={Xiong, Zuobin and Li, Wei and Li, Yingshu and Cai, Zhipeng},
  booktitle={2023 IEEE International Conference on Data Mining (ICDM)},
  pages={1439--1444},
  year={2023},
  organization={IEEE}
}

@inproceedings{yuan2023federated,
  title={Federated unlearning for on-device recommendation},
  author={Yuan, Wei and Yin, Hongzhi and Wu, Fangzhao and Zhang, Shijie and He, Tieke and Wang, Hao},
  booktitle={Proceedings of the sixteenth ACM international conference on web search and data mining},
  pages={393--401},
  year={2023}
}

@inproceedings{liu2021federaser,
  title={Federaser: Enabling efficient client-level data removal from federated learning models},
  author={Liu, Gaoyang and Ma, Xiaoqiang and Yang, Yang and Wang, Chen and Liu, Jiangchuan},
  booktitle={2021 IEEE/ACM 29th International Symposium on Quality of Service (IWQOS)},
  pages={1--10},
  year={2021},
  organization={IEEE}
}

@article{10269017,
  author={Zhao, Yian and Wang, Pengfei and Qi, Heng and Huang, Jianguo and Wei, Zongzheng and Zhang, Qiang},
  journal={IEEE Internet of Things Journal}, 
  title={Federated Unlearning With Momentum Degradation}, 
  year={2024},
  volume={11},
  number={5},
  pages={8860-8870},
  doi={10.1109/JIOT.2023.3321594}}

@misc{huynh2024fastfedultrainingfreefederatedunlearning,
      title={Fast-FedUL: A Training-Free Federated Unlearning with Provable Skew Resilience}, 
      author={Thanh Trung Huynh and Trong Bang Nguyen and Phi Le Nguyen and Thanh Tam Nguyen and Matthias Weidlich and Quoc Viet Hung Nguyen and Karl Aberer},
      year={2024},
      eprint={2405.18040},
      archivePrefix={arXiv},
      primaryClass={cs.LG},
      url={https://arxiv.org/abs/2405.18040}, 
}

@misc{jiang2024efficientcertifiedrecoverypoisoning,
      title={Towards Efficient and Certified Recovery from Poisoning Attacks in Federated Learning}, 
      author={Yu Jiang and Jiyuan Shen and Ziyao Liu and Chee Wei Tan and Kwok-Yan Lam},
      year={2024},
      eprint={2401.08216},
      archivePrefix={arXiv},
      primaryClass={cs.CR},
      url={https://arxiv.org/abs/2401.08216}, 
}

@inproceedings{cao2023fedrecover,
  title={Fedrecover: Recovering from poisoning attacks in federated learning using historical information},
  author={Cao, Xiaoyu and Jia, Jinyuan and Zhang, Zaixi and Gong, Neil Zhenqiang},
  booktitle={2023 IEEE Symposium on Security and Privacy (SP)},
  pages={1366--1383},
  year={2023},
  organization={IEEE}
}

@article{li2023subspace,
  title={Subspace based federated unlearning},
  author={Li, Guanghao and Shen, Li and Sun, Yan and Hu, Yue and Hu, Han and Tao, Dacheng},
  journal={arXiv preprint arXiv:2302.12448},
  year={2023}
}

@inproceedings{su2023asynchronous,
  title={Asynchronous federated unlearning},
  author={Su, Ningxin and Li, Baochun},
  booktitle={IEEE INFOCOM 2023-IEEE conference on computer communications},
  pages={1--10},
  year={2023},
  organization={IEEE}
}

@article{liu2021revfrf,
  title={Revfrf: Enabling cross-domain random forest training with revocable federated learning},
  author={Liu, Yang and Ma, Zhuo and Yang, Yilong and Liu, Ximeng and Ma, Jianfeng and Ren, Kui},
  journal={IEEE Transactions on Dependable and Secure Computing},
  volume={19},
  number={6},
  pages={3671--3685},
  year={2021},
  publisher={IEEE}
}

@inproceedings{liu2022right,
  title={The right to be forgotten in federated learning: An efficient realization with rapid retraining},
  author={Liu, Yi and Xu, Lei and Yuan, Xingliang and Wang, Cong and Li, Bo},
  booktitle={IEEE INFOCOM 2022-IEEE conference on computer communications},
  pages={1749--1758},
  year={2022},
  organization={IEEE}
}

@inproceedings{che2023fast,
  title={Fast federated machine unlearning with nonlinear functional theory},
  author={Che, Tianshi and Zhou, Yang and Zhang, Zijie and Lyu, Lingjuan and Liu, Ji and Yan, Da and Dou, Dejing and Huan, Jun},
  booktitle={International conference on machine learning},
  pages={4241--4268},
  year={2023},
  organization={PMLR}
}

@article{ma2022learn,
  title={Learn to forget: Machine unlearning via neuron masking},
  author={Ma, Zhuo and Liu, Yang and Liu, Ximeng and Liu, Jian and Ma, Jianfeng and Ren, Kui},
  journal={IEEE Transactions on Dependable and Secure Computing},
  volume={20},
  number={4},
  pages={3194--3207},
  year={2022},
  publisher={IEEE}
}

@misc{gu2019badnetsidentifyingvulnerabilitiesmachine,
      title={BadNets: Identifying Vulnerabilities in the Machine Learning Model Supply Chain}, 
      author={Tianyu Gu and Brendan Dolan-Gavitt and Siddharth Garg},
      year={2019},
      eprint={1708.06733},
      archivePrefix={arXiv},
      primaryClass={cs.CR},
      url={https://arxiv.org/abs/1708.06733}, 
}

@article{li2023federated,
  title={Federated unlearning via active forgetting},
  author={Li, Yuyuan and Chen, Chaochao and Zheng, Xiaolin and Zhang, Jiaming},
  journal={arXiv preprint arXiv:2307.03363},
  year={2023}
}

@inproceedings{deng2024enable,
  title={Enable the right to be forgotten with federated client unlearning in medical imaging},
  author={Deng, Zhipeng and Luo, Luyang and Chen, Hao},
  booktitle={International Conference on Medical Image Computing and Computer-Assisted Intervention},
  pages={240--250},
  year={2024},
  organization={Springer}
}

@article{wu2022federated,
  title={Federated unlearning with knowledge distillation},
  author={Wu, Chen and Zhu, Sencun and Mitra, Prasenjit},
  journal={arXiv preprint arXiv:2201.09441},
  year={2022}
}

@article{gu2024unlearning,
  title={Unlearning during learning: An efficient federated machine unlearning method},
  author={Gu, Hanlin and Zhu, Gongxi and Zhang, Jie and Zhao, Xinyuan and Han, Yuxing and Fan, Lixin and Yang, Qiang},
  journal={arXiv preprint arXiv:2405.15474},
  year={2024}
}

@article{el2022differential,
  title={Differential privacy for deep and federated learning: A survey},
  author={El Ouadrhiri, Ahmed and Abdelhadi, Ahmed},
  journal={IEEE access},
  volume={10},
  pages={22359--22380},
  year={2022},
  publisher={IEEE}
}

@misc{european_union_gdpr_2016,
    title = {General Data Protection Regulation ({GDPR})},
    shorttitle = {GDPR},
    author = {{EU}},
    type = {Regulation},
    number = {2016/679},
    year = {2016},
    month = {April},
    day = {27},
    location = {Brussels, Belgium},
    note = {Official Journal of the European Union, L 119/1},
    url = {https://eur-lex.europa.eu/legal-content/EN/TXT/?uri=CELEX:32016R}}

@INPROCEEDINGS{CaoYang2015,
  author={Yinzhi Cao and Junfeng Yang},
  booktitle={2015 IEEE Symposium on Security and Privacy}, 
  title={Towards Making Systems Forget with Machine Unlearning}, 
  year={2015},
  volume={},
  number={},
  pages={463-480},
  doi={10.1109/SP.2015.35}}
\bibliographystyle{iclr2027_conference}

\newpage
\appendix

\section{Other Related Works}\label{appendix:A}

\textbf{Machine Unlearning}:
Machine unlearning seeks to remove the influence of specific data from trained models, a task that has grown increasingly complex with the rise of deep learning. Early efforts established a mathematical framework inspired by differential privacy, achieving success in convex optimization  problems\citep{sekhari2021rememberwantforgetalgorithms}. However, modern deep learning models, with their non-convex objectives, are prone to memorization, retaining specific training data within their parameters \citep{li2024machine}. This persistence of sensitive or malicious data complicates unlearning, as standard removal techniques often fail to fully erase these traces. Further complicating the task, research has shown that models trained on non-overlapping datasets can converge to nearly identical weights, suggesting that a specific parameter state does not guarantee effective unlearning \citep{thudi2022unrolling}.\par
Two primary approaches have emerged: exact unlearning \citep{cao2015towards} and approximate unlearning  \citep{nguyen2020variational}. Exact unlearning involves retraining the model from scratch using only the remaining data, a computationally expensive process impractical for large-scale models \citep{cao2015towards, thudi2022unrolling}. In contrast, approximate unlearning adjusts the trained model to approximate the outcome of retraining, requiring robust theoretical guarantees to ensure reliability \citep{guo2019certified}. The non-convex nature of deep neural networks poses a significant hurdle, with heuristic methods yielding inconsistent results across benchmarks \citep{li2024machine}.\par
To address these challenges, researchers have explored data-driven unlearning techniques. The OracleMatching method enhances stability during unlearning \citep{georgiev2024attributetodeletemachineunlearningdatamodel}, while sparsity-regularized fine-tuning reduces computational overhead. Bayesian and variational inference provide probabilistic frameworks to estimate the impact of forgotten data \citep{nguyen2020variational}. Data attribution techniques identify and remove targeted data influences but risk leaking information in federated settings. A critical issue, the missing targets problem, arises in fine-tuning-based methods. When unlearning a data point, gradient ascent is applied to the forget set, while gradient descent preserves performance on the retain set. Without knowing the ideal loss—equivalent to a model trained only on retained data—the process may overshoot or undershoot, destabilizing the model \citep{hayes2024}.\par
Recent advances leverage f-divergences, such as Jensen-Shannon divergence, to design loss functions that mitigate these issues, ensuring more stable unlearning \citep{bonato2025retain, jeon2024information, rangel2024learning}. These tools, previously used to validate machine learning tasks \citep{aminian2024robustsemisupervisedlearningfdivergence, roulet2025lossfunctionsoperatorsgenerated, novello2024fdivergencebasedclassificationuse}, show particular promise for unlearning in large language models \citep{wang2024llm}. Together, these efforts reflect a dynamic field, navigating the delicate balance of efficiency, privacy, and model integrity in the pursuit of effective machine unlearning.\par

\textbf{Beyond Accumulator:} In the domain of active federated unlearning, \citep{jin2024forgettablefederatedlinearlearning} proposed Forgettable Federated Linear Learning with Certified Data Removal, enabling provable data removal in linear models by exploiting their analytical properties. While computationally efficient and privacy-preserving in theory, the method holds a dataset on server for calculating Hessian approximation. Several early works, such as \citep{xiong2023exact} and \citep{xiong2024appro}, addressed unlearning in convex optimization settings. However, given that most modern deep learning systems involve non-convex objectives, the conclusions drawn from these convex approaches are of limited utility in the most ML frameworks. 
 
 To address unlearning in more complex models, \citep{wang2024goldfishefficientfederatedunlearning} introduced Goldfish, a framework that removes client influence without full retraining. It incorporates a novel loss balancing retained accuracy, removal bias, and confidence. However, the method needs a portion of data on the server side. In \citep{liu2022right}, the model scrubbing technique is applied locally on the target client to unlearn specific portions of data. This process involves Hessian matrix computations, with improved computational efficiency achieved by approximating the diagonal of the empirical Fisher Information Matrix (FIM).

\citep{che2023fast} introduces FFMU, an efficient federated unlearning algorithm that leverages nonlinear functional analysis specifically, the Nemytskii operator to transform local models into smooth, quantized representations, enabling the server to aggregate them into a global model with certified unlearning guarantees. \citep{deng2024enable}, presents a novel Federated Client Unlearning framework to address the right to be forgotten in medical imaging, utilizing Model Contrastive Unlearning (MCU) to align the unlearned model with a downgraded model at the feature level and Frequency Guided Memory Preservation (FGMP) to preserve general knowledge. \citep{wu2022federated} proposes a novel federated unlearning method that removes a client's contribution from a trained global model by subtracting its historical updates and recovering performance through knowledge distillation on the server side. \citep{10269017} approach splits unlearning into (i) knowledge erasure via momentum degradation and (ii) memory guidance via guided fine-tuning to restore discriminability on retained data. 

\section{Algorithm}
\label{appendix:Alg}

The \method algorithm is presented in
Algorithm~\ref{alg:fed-f-scrub-gradient-local}.

The \method algorithm introduces several hyperparameters to control the
dynamics of federated unlearning. Specifically, it uses $T$ max--min
unlearning iterations and $T_{\mathrm{post}}$ post-training iterations
to refine the final model. Learning rates $\eta_{\max}$ and
$\eta_{\min}$ are used during the maximization (forgetting) and
minimization (retention and post-training) phases, respectively. The
number of local training epochs for each client is controlled by
$E_{\max}$, $E_{\min}$, and $E_{\mathrm{post}}$, corresponding to the
forgetting, retention, and post-training phases.

\begin{algorithm}[tb]
\caption{\method: Federated Unlearning via Divergence Optimization
(Gradient Version with Local Rounds)}
\label{alg:fed-f-scrub-gradient-local}
\begin{algorithmic}[1]

\REQUIRE Initial model $\theta^0$; iterations $T$, $T_{\mathrm{post}}$;
loss weights $\alpha,\gamma$; learning rates
$\eta_{\max},\eta_{\min}$; local epochs
$E_{\max},E_{\min},E_{\mathrm{post}}$;
client-local forget and retain sets
$\mathcal{S}_F^{(c)},\mathcal{S}_R^{(c)}$

\FUNCTION{LocalUpdate($\theta$, $\theta_{\mathrm{ref}}$, $\eta$, $\mathcal{L}$, $E$)}
    \FOR{$e=1$ \textbf{to} $E$}
        \STATE $g \gets \nabla_\theta
        \mathcal{L}(\theta;\theta_{\mathrm{ref}})$
        \STATE $\theta \gets \theta-\eta g$
    \ENDFOR
    \STATE \textbf{return} $\theta$
\ENDFUNCTION

\STATE $\theta_{\mathrm{ref}} \gets \theta^0$
\COMMENT{fixed (evaluation mode) throughout unlearning and post-training }

\STATE $\mathcal{C}_F \gets
    \{c\in[C]:n_F^{(c)}>0\}$,\quad
    $\mathcal{C}_R \gets
    \{c\in[C]:n_R^{(c)}>0\}$

\STATE $n_F \gets
    \sum_{c\in\mathcal{C}_F}n_F^{(c)}$,\quad
    $n_R \gets
    \sum_{c\in\mathcal{C}_R}n_R^{(c)}$

\FOR{$t=0$ \textbf{to} $T-1$}

    \FORALL{clients $c\in\mathcal{C}_F$ in parallel}
        \STATE $\theta_{\max}^{(c)} \gets
        \textsc{LocalUpdate}(
        \theta^t,
        \theta_{\mathrm{ref}},
        -\eta_{\max},
        \mathcal{L}_{\max}^{(c)},
        E_{\max})$
    \ENDFOR

    \STATE $\theta^{t+\frac{1}{2}} \gets
    \sum_{c\in\mathcal{C}_F}
    \frac{n_F^{(c)}}{n_F}
    \theta_{\max}^{(c)}$

    \FORALL{clients $c\in\mathcal{C}_R$ in parallel}
        \STATE $\theta_{\min}^{(c)} \gets
        \textsc{LocalUpdate}(
        \theta^{t+\frac{1}{2}},
        \theta_{\mathrm{ref}},
        \eta_{\min},
        \mathcal{L}_{\min}^{(c)}(\alpha,\gamma),
        E_{\min})$
    \ENDFOR

    \STATE $\theta^{t+1} \gets
    \sum_{c\in\mathcal{C}_R}
    \frac{n_R^{(c)}}{n_R}
    \theta_{\min}^{(c)}$

\ENDFOR

\FOR{$\tau=0$ \textbf{to} $T_{\mathrm{post}}-1$}

    \FORALL{clients $c\in\mathcal{C}_R$ in parallel}
        \STATE $\theta_{\mathrm{post}}^{(c)} \gets
        \textsc{LocalUpdate}(
        \theta^{T+\tau},
        \theta_{\mathrm{ref}},
        \eta_{\min},
        \mathcal{L}_{\min}^{(c)}(\alpha,\gamma),
        E_{\mathrm{post}})$
    \ENDFOR

    \STATE $\theta^{T+\tau+1} \gets
    \sum_{c\in\mathcal{C}_R}
    \frac{n_R^{(c)}}{n_R}
    \theta_{\mathrm{post}}^{(c)}$

\ENDFOR

\STATE $\theta_U \gets \theta^{T+T_{\mathrm{post}}}$
\STATE \textbf{return} $\theta_U$

\end{algorithmic}
\end{algorithm}

\textbf{Federated Updates (Gradient Version):}
At iteration $t$, the server broadcasts the current global student
model $\theta^t$. Clients first perform the forgetting update on their
local forget sets:
\begin{align}
g_{\max}^{(c)}
&=
\nabla_\theta
\mathcal{L}_{\max}^{(c)}
\left(
S_F^{(c)},
\theta^t,
\theta_{\mathrm{ref}}
\right),
\\
\theta_{\max}^{(c)}
&=
\theta^t
+
\eta_{\max}
g_{\max}^{(c)},
\\
\theta^{t+\frac{1}{2}}
&=
\sum_{c\in\mathcal{C}_F}
\frac{n_F^{(c)}}{n_F}
\theta_{\max}^{(c)}.
\end{align}

The aggregated model $\theta^{t+\frac{1}{2}}$ is then broadcast to the
clients participating in the retention phase:
\begin{align}
g_{\min}^{(c)}
&=
\nabla_\theta
\mathcal{L}_{\min}^{(c)}
\left(
S_R^{(c)},
\theta^{t+\frac{1}{2}},
\theta_{\mathrm{ref}}
\right),
\\
\theta_{\min}^{(c)}
&=
\theta^{t+\frac{1}{2}}
-
\eta_{\min}
g_{\min}^{(c)},
\\
\theta^{t+1}
&=
\sum_{c\in\mathcal{C}_R}
\frac{n_R^{(c)}}{n_R}
\theta_{\min}^{(c)}.
\end{align}

Following Algorithm~\ref{alg:fed-f-scrub-gradient-local}, the reference
(teacher) model is fixed to the original pretrained model throughout
both the unlearning and post-training phases:
\begin{equation}
\theta_{\mathrm{ref}}=\theta^0.
\end{equation}
The student model evolves across the unlearning iterations, whereas
$\theta_{\mathrm{ref}}$ remains unchanged.

\textbf{Post-training Minimization (Gradient Version):}
After $T$ max--min unlearning iterations, the model undergoes
$T_{\mathrm{post}}$ minimization-only iterations:
\begin{equation}
\begin{split}
g_{\min}^{(c)}
&=
\nabla_\theta
\mathcal{L}_{\min}^{(c)}
\left(
S_R^{(c)},
\theta^{T+\tau},
\theta_{\mathrm{ref}}
\right),
\\
\theta_{\mathrm{post}}^{(c)}
&=
\theta^{T+\tau}
-
\eta_{\min}
g_{\min}^{(c)},
\\
\theta^{T+\tau+1}
&=
\sum_{c\in\mathcal{C}_R}
\frac{n_R^{(c)}}{n_R}
\theta_{\mathrm{post}}^{(c)},
\qquad
\tau=0,\ldots,T_{\mathrm{post}}-1.
\end{split}
\end{equation}

The final unlearned model is therefore
\begin{equation}
\theta_U
=
\theta^{T+T_{\mathrm{post}}}.
\end{equation}

For client-level unlearning, the target client $k$ participates only in
the forgetting phase and is excluded from the retained-data phase.
Accordingly,
\begin{equation}
n_R^{(k)}=0.
\end{equation}
Conversely, any client that does not participate in the forgetting phase
has
\begin{equation}
n_F^{(c)}=0.
\end{equation}

Note that when the student and reference models have identical parameters, 
the first maximization gradient may be zero if their predictive distributions 
are identical. In our implementation, the reference model is fixed and 
evaluated in evaluation mode, while the student is evaluated in training mode 
during local updates. Consequently, training-mode-dependent components can 
produce different predictive distributions even when the two models initially 
share identical parameters, and the initial forget gradient therefore need 
not be zero.

\section{Motivations for JS-divergence and \texorpdfstring{$\chi^2$}{chi-square}-divergence}\label{app:motive-js-chi}
 
\textbf{$f$-\textbf{divergence}:} The $f$-divergences are information measures that generalize various divergences, such as Kullback-Leibler (KL) divergence, through the use of a convex generator function $f$. Given two discrete distributions $P = \{p_i\}_{i=1}^{k}$ and $Q = \{q_i\}_{i=1}^{k}$, the $f$-divergence between them is defined as:
\begin{equation}
D_f(P \parallel Q) := \sum_{i=1}^{k} q_i f\left( \frac{p_i}{q_i} \right)
\end{equation}
where $f: (0, \infty) \to \mathbb{R}$ is a convex function with the property that $f(1) = 0$. In this work, we focus on KL-divergence, JS-divergence and $\chi^2$-divergence (Table \ref{tab:divergences}. Therefore, we consider $D_f\in\{D_{\mathrm{KL}},D_{\mathrm{JS}}, D_{\mathrm{\chi^2}}\}$. This definition implies that $D_f(P \parallel Q) = 0$ if and only if $P = Q$. By choosing different forms for $f$, we obtain different types of divergences. For example, when $f(t) = t \log(t)$, we get the Kullback-Leibler (KL) divergence, which measures the difference between two probability distributions.

\begin{table}[h]
\caption{Divergences and their corresponding generator functions} \label{tab:divergences}
\begin{center}
\begin{tabular}{ll}
\toprule
\textbf{Divergence} & \textbf{Generator Function} \\
\midrule \\
KL-divergence & $t \log t$ \\
$\chi^2$-divergence & $(1 - t)^2$ \\
JS-divergence & $\frac{1}{2}(t \log \left( \frac{2t}{1+t} \right) + \log \left( \frac{2}{1+t} \right))$ \\
\bottomrule
\end{tabular}
\end{center}
\end{table}

In this section, we study some motivations behind choosing JS-divergence and $\chi^2$ divergence. These information measures offer several advantages over KL divergence, particularly in applications involving generative modeling and robust regularization.

JS-divergence is widely used as a loss function in Generative Adversarial Networks (GANs) due to its symmetric and bounded nature, which provides a stable measure of similarity between distributions \citep{goodfellow2014generativeadversarialnetworks}. Unlike KL divergence, which can diverge to infinity when the two distributions have disjoint supports, JS-divergence remains finite and well-behaved, making it particularly effective for comparing empirical distributions (\citep{nowozin2016fgantraininggenerativeneural}). This property is especially beneficial in our context, as it helps mitigate overshoot and undershoot problems, particularly in scenarios where exact loss values for removed data points are unavailable.

On the other hand, $\chi^2$ divergence emphasizes large discrepancies due to its squared difference term, making it particularly useful in outlier detection and robust learning frameworks \citep{reid2009informationdivergenceriskbinary}. Regularizing with $\chi^2$ divergence can also help prevent models from becoming overly biased toward majority classes by strongly penalizing large probability gaps \citep{duchi2020learningmodelsuniformperformance}. This property makes it particularly effective in imbalanced learning scenarios, where standard loss functions may fail to capture significant disparities between class distributions.

Thus, by leveraging JS-divergence for stable probability comparisons and $\chi^2$ divergence for strong regularization and outlier sensitivity, we can achieve a more robust and balanced learning framework compared to using KL divergence alone.

Building on this, we modify our loss functions and introduce  \method, where we select different $f$-divergences for the retain set and the forget set. Each divergence term, $d(x_r; \theta^u)$ and $d(x_f; \theta^u)$, can be chosen from JS, KL, or $\chi^2$ divergences \citep{Nguyen_2010}.

\section{Implementation Details}
\label{sec:implementation-details}

\subsection{Federated Data Partitioning}

We evaluate \emph{f-FUM} on CIFAR-10, CIFAR-100, MNIST, and
FashionMNIST under federations containing either \(N=5\) or \(N=10\)
clients. The training data are partitioned across clients using a
class-balanced IID partition: samples from every class are divided as
evenly as possible among the clients. Consequently, client dataset sizes
are approximately equal, up to rounding when a class size is not
divisible by \(N\). All available clients participate in every
federated round.

Within each client partition, the data are split in a class-stratified
manner into a local optimization set and a held-out validation set.
The validation data are clean, are not used for local optimization, and
are disjoint from the forget, retained-training, and official test sets.
The same saved partition and forget-set indices are used by all compared
methods for a given experimental setting.

\subsection{Models and Pretraining Protocol}

We use ResNet-18 for CIFAR-100, the reduced ResNet-18 architecture for
CIFAR-10, LeNet-5 for MNIST, and the lightweight two-convolutional-layer
network for FashionMNIST. The federated system is implemented using
Flower~\citep{beutel2022flowerfriendlyfederatedlearning}. Flower executes
the participating clients in parallel: all ten clients are run in
parallel for \(N=10\), and all five clients are run in parallel for
\(N=5\).

Client-side optimization is performed using Adam with
\(\beta_1=0.9\) and \(\beta_2=0.999\). We use a batch size of \(512\)
throughout the reported experiments. Pretraining continues until the
training loss has converged. In particular, CIFAR-10 is pretrained for
\(200\) global communication rounds, while MNIST and FashionMNIST are
pretrained for \(50\) rounds.  CIFAR-100 is Pretrain on 300 rounds but with two local epoch rather than one.

For each experimental configuration, every unlearning method starts
from exactly the same pretrained global checkpoint. All methods also
receive the same deletion request and exactly the same forget-set
indices. Thus, differences between methods cannot be attributed to
different pretrained models or different sampled deletion requests.

\begin{table*}[t]
\centering
\caption{Dataset-specific model architectures and federated pretraining settings.
The reported pretraining rounds correspond to budgets for which all observed
pretraining runs had reached convergence. Unless otherwise specified, each client
performs one local epoch per communication round; for CIFAR-100, each client
performs two local epochs per round.}
\label{tab:pretraining-implementation}

\small
\begin{tabular}{lccc}
\toprule
\textbf{Dataset} &
\textbf{Model} &
\textbf{Local epochs} &
\textbf{Pretraining rounds} \\
\midrule

CIFAR-10
& Reduced ResNet-18
& \(1\)
& \(200\) \\

CIFAR-100
& ResNet-18
& \(2\)
& \(300\) \\

MNIST
& LeNet-5
& \(1\)
& \(50\) \\

FashionMNIST
& Lightweight CNN
& \(1\)
& \(50\) \\

\bottomrule
\end{tabular}
\end{table*}

\subsection{Construction of Unlearning Requests}

We consider privacy-oriented deletion, label-confusion removal, and
backdoor removal.

\paragraph{Privacy.}
The forget set consists of clean training examples selected for deletion.
No corruption or label noise is added in this setting. This avoids
confounding unlearning with the privacy benefits that may arise from
training on noisy data. In data-level unlearning, the requested fraction
is removed from every affected client's local data. In client-level
unlearning, the contribution of the target client is removed.

\paragraph{Label confusion.}
In the \emph{confuse} scenario, labels are deterministically swapped
within paired classes, for example \(1\leftrightarrow2\) and
\(3\leftrightarrow4\). The same class-pair mapping is used throughout a
given experiment. During evaluation, the model is assessed on both the
original clean labels and the confused-label view of the forget set.

\paragraph{Backdoor.}
Following~\citet{halimi2023federatedunlearningefficientlyerase}, a fixed
fraction of the target client's images is modified by inserting a
\(3\times3\) pixel-pattern trigger. The triggered examples are relabeled
to target class \(0\). Backdoor unlearning is evaluated using both clean
test accuracy and attack success rate (ASR), where lower ASR is
better.

For every method and repetition, the underlying pretrained checkpoint
and deletion request remain fixed. Only stochastic operations performed
during unlearning, such as minibatch ordering and applicable client
sampling, vary between repetitions.

\subsection{\texorpdfstring{\(f\)-FUM}{f-FUM} Optimization}

The maximization phase increases the predictive divergence between the
student and reference models on the forget set. The minimization phase
anchors the student to the reference model on retained data while also
minimizing the supervised retained-data loss. The corresponding local
retained-data objective is weighted by \(\alpha\) and \(\gamma\):
\(\alpha\) controls teacher--student alignment on retained samples,
whereas \(\gamma\) controls the supervised retained-data loss.

We select
\[
    \alpha=0.5,
    \qquad
    \gamma=1.
\]
These values were selected using centralized, single-client
cross-validation. The selected values were then frozen and transferred
to the \(N=5\) and \(N=10\) federated experiments. This strategy avoids
requiring the complete federation to participate repeatedly during
hyperparameter selection.

The KL, Jensen--Shannon, and \(\chi^2\) variants use the same data
partitions, initialization, communication budget, optimizer protocol,
and checkpoint-selection rule. The divergence is the experimental
factor being changed; it is not accompanied by a different deletion
request or checkpoint-selection criterion.

For CIFAR-10, we perform \(T=5\) max-min iterations,
followed by \(T_{\mathrm{post}}=10\) minimization-only recovery phases.
Because each max-min iteration contains one maximization synchronization
and one minimization synchronization, the total cost is
\[
    B = 2T+T_{\mathrm{post}}
      = 2(5)+10=20
\]
synchronization phases.

For MNIST and FashionMNIST, we use
\(T=3\) and \(T_{\mathrm{post}}=4\), giving
\[
    B=2(3)+4=10
\]
synchronization phases.

For CIFAR-100, we use \(T=2\) max-min iterations.
These iterations are followed by \(T_{\mathrm{post}}=26\)
retained-data recovery phases. The total communication budget is
therefore
\[
    B = 2T+T_{\mathrm{post}}
      = 2(2)+26
      = 30
\]
synchronization phases.

The original pretrained model is used as the teacher/reference model
during unlearning. This fixes the target distribution and ensures that
all divergence variants are compared against the same reference.

\subsection{Baseline Fairness and Communication Budget}

We compare against NoT, Halimi, and MoDE. For each fixed dataset and
federation size, all approximate-unlearning methods receive the same
predefined synchronization-phase budget \(B\). A synchronization phase
consists of
\[
 \text{server broadcast}
 \rightarrow
 \text{local client computation}
 \rightarrow
 \text{client upload}
 \rightarrow
 \text{server aggregation}.
\]
Thus, the comparison controls the number of opportunities each method
has to communicate with the federation.

For \(f\)-FUM, NoT, Halimi, and MoDE, respectively, the phase counts can
be written as
\[
\begin{aligned}
C_{f\text{-FUM}} &= 2T + T_{\mathrm{post}},\\
C_{\mathrm{NoT}} &= T_{\mathrm{neg}} + T_{\mathrm{fine}},\\
C_{\mathrm{Halimi}} &= T_{\mathrm{ascent}} + T_{\mathrm{fine}},\\
C_{\mathrm{MoDE}} &= T_{\mathrm{deg}} + T_{\mathrm{guid}}.
\end{aligned}
\]
Here, $T_{\mathrm{neg}}$ and $T_{\mathrm{fine}}$ refer to the weight-negation and fine-tuning phases of NoT \citep{khalil2025notfederatedunlearningweight}, $T_{\mathrm{ascent}}$ and $T_{\mathrm{fine}}$ to the gradient-ascent and fine-tuning phases of Halimi \citep{halimi2023federatedunlearningefficientlyerase}, and $T_{\mathrm{deg}}$ and $T_{\mathrm{guid}}$ to the degradation and guidance phases of MoDE \citep{10269017}. The experimental protocol enforces
\[
C_{f\text{-FUM}}
=
C_{\mathrm{NoT}}
=
C_{\mathrm{Halimi}}
=
C_{\mathrm{MoDE}}
=
B.
\]

Exact retraining is different: it repeats the complete federated
pretraining schedule on the retained dataset. Consequently, approximate
unlearning requires substantially fewer synchronization phases than
exact retraining.

Equal synchronization-phase counts do not necessarily imply equal
transmitted bytes, local computation, or wall-clock time. Because these
quantities were not measured consistently for every baseline, we do not
claim a head-to-head byte-level or wall-clock advantage over NoT,
Halimi, or MoDE. The communication reduction above is specifically
relative to exact retraining.

\subsection{Model Selection, Repetitions, and Metrics}

Every method is evaluated within its fixed communication budget. Among
the checkpoints produced within that budget, we select the checkpoint
with the highest accuracy on the held-out validation set. All reported
metrics including clean accuracy, forget-set accuracy, confusion
accuracy, ASR, the standard MIA score, and Yeom's attack are evaluated
using that same selected checkpoint. Neither test-set performance nor
MIA performance is used for checkpoint selection.

The revised principal experiments use three independent unlearning
seeds,
$
    \{7,17,101\}.
$
We report the mean and standard deviation over these three repetitions.
The pretrained global checkpoint, client partition, and forget-set
indices remain fixed. The seeds vary stochastic operations during
unlearning and MIA evaluation. Therefore, the reported variability
measures uncertainty conditional on a fixed pretrained model and fixed
deletion request.

\subsection{Runtime}

All experiments were run on a machine equipped with one NVIDIA H100
GPU, ten CPU cores, and \(512\) GB of RAM. With Flower executing clients
in parallel, CIFAR-10 pretraining took approximately \(2.30\) hours and
unlearning approximately \(20\) minutes. For MNIST and FashionMNIST,
pretraining took approximately \(25\) minutes and unlearning
approximately \(2.5\) minutes.
\paragraph{Why $f$-divergences and why Jensen--Shannon for maximization.}
We formulate the maximization step using an $f$-divergence, and in particular the Jensen--Shannon (JS) divergence, for stability reasons.
Unlike KL divergence, Jensen--Shannon divergence is bounded above by
$\log 2$. Thus, the forget-set JS objective cannot increase without
bound. This motivates its use as a potentially more conservative
forgetting objective, and is consistent with the favorable
forgetting-utility trade-offs observed in some of our experiments.
Boundedness of the divergence alone does not guarantee bounded
parameter updates or prevent model drift; the optimization behavior
also depends on model gradients, learning rates, and the number of
updates.
Empirically, this stabilizes the alternating optimization and reduces the risk of irreversible drift that would otherwise degrade recovery.


\section{Additional Results}\label{App:E}

\begin{table*}[t]
\centering

\begin{minipage}[t]{0.68\textwidth}
\vspace{0pt}
\centering

\captionof{table}{
Comparison of data-level unlearning performance under backdoor and
confused-data scenarios for $N=10$ clients.
The setting $N/r$ denotes $N$ clients with $r\%$ of poisoned data removed
from each client. Higher accuracy is better.
}
\label{tab:data-level-backdoor-confused-n10}

\setlength{\tabcolsep}{4.0pt}
\renewcommand{\arraystretch}{1.10}

\resizebox{\linewidth}{!}{%
\begin{tabular}{l c cc cc cc}
\toprule

\multirow{3}{*}{\textbf{Method}}
& \multirow{3}{*}{\textbf{Setting}}
& \multicolumn{4}{c}{\textbf{Backdoor}}
& \multicolumn{2}{c}{\textbf{Confused}} \\

\cmidrule(lr){3-6}
\cmidrule(lr){7-8}

&
& \multicolumn{2}{c}{\textbf{CIFAR-10}}
& \multicolumn{2}{c}{\textbf{Fashion-MNIST}}
& \multicolumn{2}{c}{\textbf{CIFAR-10}} \\

\cmidrule(lr){3-4}
\cmidrule(lr){5-6}
\cmidrule(lr){7-8}

&
& \textbf{Pre.} & \textbf{Unl.}
& \textbf{Pre.} & \textbf{Unl.}
& \textbf{Pre.} & \textbf{Unl.} \\

\midrule


Ours (KL)
& \multirow{5}{*}{10/6}
& \multirow{5}{*}{28.42} & 52.62
& \multirow{5}{*}{65.93} & 77.10
& \multirow{5}{*}{60.44} & 67.84 \\

Ours (JS)
&
& & \textbf{73.74}
& & \textbf{79.90}
& & \textbf{70.72} \\

Ours ($\chi^2$)
&
& & 44.78
& & 76.25
& & 61.30 \\

NoT
&
& & 43.00
& & 26.67
& & 68.00 \\

Halimi
&
& & 43.00
& & 77.82
& & 68.00 \\

\cmidrule(lr){2-8}


Ours (KL)
& \multirow{5}{*}{10/10}
& \multirow{5}{*}{38.42} & 52.74
& \multirow{5}{*}{41.98} & 49.20
& \multirow{5}{*}{57.10} & 64.06 \\

Ours (JS)
&
& & \textbf{72.32}
& & 70.48
& & \textbf{67.26} \\

Ours ($\chi^2$)
&
& & 52.34
& & 48.17
& & 61.40 \\

NoT
&
& & 42.00
& & 32.13
& & 65.00 \\

Halimi
&
& & 42.00
& & \textbf{79.58}
& & 65.00 \\

\bottomrule
\end{tabular}%
}

\end{minipage}
\hfill
\begin{minipage}[t]{0.30\textwidth}
\vspace{0pt}
\centering

\captionof{table}{
Data-level unlearning performance under the confused-data scenario
for $N=5$ clients. The setting $N/r$ denotes $N$ clients with $r\%$
of poisoned data removed from each client. Higher accuracy is better.
}
\label{tab:data-level-confused-n5}

\setlength{\tabcolsep}{5pt}
\renewcommand{\arraystretch}{1.10}

\resizebox{\linewidth}{!}{%
\begin{tabular}{l c cc}
\toprule

\multirow{2}{*}{\textbf{Method}}
& \multirow{2}{*}{\textbf{Setting}}
& \multicolumn{2}{c}{\textbf{CIFAR-10}} \\

\cmidrule(lr){3-4}

&
& \textbf{Pre.} & \textbf{Unl.} \\

\midrule


Ours (KL)
& \multirow{5}{*}{5/6}
& \multirow{5}{*}{56.86} & \textbf{74.20} \\

Ours (JS)
&
& & 69.80 \\

Ours ($\chi^2$)
&
& & 62.22 \\

NoT
&
& & 63.00 \\

Halimi
&
& & 73.00 \\

\cmidrule(lr){2-4}


Ours (KL)
& \multirow{5}{*}{5/10}
& \multirow{5}{*}{54.38} & 63.24 \\

Ours (JS)
&
& & \textbf{69.70} \\

Ours ($\chi^2$)
&
& & 60.10 \\

NoT
&
& & 64.00 \\

Halimi
&
& & 64.00 \\

\bottomrule
\end{tabular}%
}

\end{minipage}

\end{table*}
\subsection{Data-Level Experiments}
In the tables \ref{tab:data-level-backdoor-confused-n10}- \ref{tab:unlearning-data-privacy-6-10percent} we see the result of our model for the harder unlearning setup; there are couple things to mention; 1) Some attacks like Confuse attack for MNIST dataset under 10 clients and 10\% data removal would not let the model converge and the accuracy and the model are not being practical; 2) on the other hand; there exist some situation where the accuracy is too low but under our unlearning setup we achieve more than 90\% accuracy. (see table \ref{tab:data-level-backdoor-confused}); This leads to a main question of how our model is robust through the  model architecture and unlearning situation.
\subsection{Network Architecture Sensitivity}
On CIFAR-10 with ResNet-18, all methods face challenges due to the dataset's inherent difficulty and the model's capacity. Shallow architectures present different challenges. This architectural sensitivity likely stems from shallow networks' limited capacity to independently represent both corrupted and clean patterns, making the unlearning objective more critical for success. Our divergence-based approach appears to be better suited to guide this limited capacity toward the desired solution. (e.g. See The NoT performance in the Table \ref{tab:data-level-backdoor-confused-n10} and Table \ref{tab:data-level-confused-n5})

\subsection{The Privacy-Performance dilemma}

\begin{table*}[t]
\caption{Data-level unlearning results under the privacy scenario for CIFAR-10,
Fashion-MNIST, and MNIST. The setting $N/r$ denotes $N$ clients with $r\%$
of local clean data removed from each client. Accuracy is higher-is-better,
while MIA is best when closest to 50.}
\label{tab:unlearning-data-privacy-6-10percent}
\centering
\setlength{\tabcolsep}{4.0pt}
\renewcommand{\arraystretch}{1.08}

\resizebox{\textwidth}{!}{%
\begin{tabular}{c l ccc ccc ccc}
\toprule

\multirow{2}{*}{\textbf{Setting}}
& \multirow{2}{*}{\textbf{Method}}
& \multicolumn{3}{c}{\textbf{CIFAR-10}}
& \multicolumn{3}{c}{\textbf{Fashion-MNIST}}
& \multicolumn{3}{c}{\textbf{MNIST}} \\

\cmidrule(lr){3-5}
\cmidrule(lr){6-8}
\cmidrule(lr){9-11}

&
& \textbf{Pre. Acc.}
& \textbf{Unl. Acc.}
& \textbf{MIA}
& \textbf{Pre. Acc.}
& \textbf{Unl. Acc.}
& \textbf{MIA}
& \textbf{Pre. Acc.}
& \textbf{Unl. Acc.}
& \textbf{MIA} \\

\midrule


\multirow{5}{*}{10/6}
& Ours (KL)
& \multirow{5}{*}{79.00}
& 78.10
& 61.44
& \multirow{5}{*}{99.13}
& 77.65
& 49.03
& \multirow{5}{*}{99.04}
& 78.68
& 61.04 \\

& Ours (JS)
&
& 77.92
& 61.33
&
& 78.17
& 48.94
&
& 78.66
& 59.74 \\

& Ours ($\chi^2$)
&
& 77.82
& 62.00
&
& 77.23
& 49.02
&
& 78.52
& 62.11 \\

& NoT
&
& 74.14
& 51.77
&
& 73.38
& 49.13
&
& 73.44
& 52.11 \\

& Halimi
&
& 78.52
& 61.66
&
& 77.43
& 48.97
&
& 78.44
& 62.48 \\

\cmidrule(lr){1-11}


\multirow{5}{*}{10/10}
& Ours (KL)
& \multirow{5}{*}{79.00}
& 78.84
& 59.88
& \multirow{5}{*}{99.20}
& 79.40
& 48.57
& \multirow{5}{*}{99.13}
& 78.34
& 59.78 \\

& Ours (JS)
&
& 78.64
& 59.33
&
& 80.10
& 48.96
&
& 78.80
& 58.91 \\

& Ours ($\chi^2$)
&
& 78.08
& 60.22
&
& 78.87
& 48.33
&
& 78.12
& 60.44 \\

& NoT
&
& 70.14
& 51.42
&
& 38.52
& 48.52
&
& 70.32
& 51.42 \\

& Halimi
&
& 78.28
& 61.06
&
& 79.18
& 48.48
&
& 78.86
& 60.44 \\

\midrule


\multirow{5}{*}{5/6}
& Ours (KL)
& \multirow{5}{*}{80.00}
& 80.00
& 58.96
& \multirow{5}{*}{99.03}
& 81.43
& 50.91
& \multirow{5}{*}{99.01}
& 80.06
& 57.89 \\

& Ours (JS)
&
& 78.32
& 59.29
&
& 80.77
& 50.50
&
& 78.40
& 59.593 \\

& Ours ($\chi^2$)
&
& 79.66
& 59.67
&
& 80.18
& 51.31
&
& 77.54
& 57.74 \\

& NoT
&
& 74.82
& 54.00
&
& 44.68
& 51.00
&
& 75.02
& 54.19 \\

& Halimi
&
& 80.62
& 58.55
&
& 77.07
& 51.34
&
& 80.20
& 58.30 \\

\cmidrule(lr){1-11}


\multirow{5}{*}{5/10}
& Ours (KL)
& \multirow{5}{*}{80.00}
& 77.64
& 61.24
& \multirow{5}{*}{99.08}
& 78.73
& 49.41
& \multirow{5}{*}{99.08}
& 78.48
& 61.53 \\

& Ours (JS)
&
& 78.32
& 60.86
&
& 49.56
& 49.56
&
& 78.26
& 60.89 \\

& Ours ($\chi^2$)
&
& 77.90
& 61.62
&
& 78.07
& 49.32
&
& 77.68
& 61.78 \\

& NoT
&
& 72.74
& 52.68
&
& 72.03
& 49.17
&
& 71.44
& 52.64 \\

& Halimi
&
& 77.70
& 62.26
&
& 78.30
& 49.48
&
& 78.04
& 60.71 \\

\bottomrule
\end{tabular}%
}
\end{table*}

The privacy-oriented unlearning scenarios revealed in Tables \ref{tab:cifar100-combined}, \ref{tab:privacy-unlearning-1}, and \ref{tab:privacy-unlearning-2}, and \ref{tab:unlearning-data-privacy-6-10percent} expose a fundamental tension inherent to federated unlearning: achieving strong privacy guarantees (measured by MIA scores approaching 50\%) often comes at the cost of model utility. Unlike robustness-oriented scenarios where removing corrupted data improves performance, privacy-oriented unlearning inherently involves sacrificing useful information, making this trade-off unavoidable. Our empirical analysis across Tables 4, 7, and 10 reveals that this dilemma manifests differently depending on task complexity and federation structure: on simple tasks like MNIST, all divergence variants achieve near-perfect privacy (MIA $\approx$ 50\%) with minimal utility loss, while complex tasks like CIFAR-10 expose a clear Pareto frontier where practitioners must choose between aggressive forgetting (KL-JS achieving 55.20\% MIA but 63.16\% accuracy) and utility preservation (KL-$\chi^2$ maintaining 77.66\% accuracy but 64.60\% MIA). Notably, we observe that fewer clients ($N=5$) generally yield better outcomes in both dimensions compared to more fragmented settings ($N=10$), and that fine-grained data-level unlearning (2-10\% per client) persistently struggles to achieve ideal privacy metrics even when utility is well-preserved, suggesting fundamental limitations in current federated unlearning approaches for distributed deletion requests.

\section{Case study: Single Client Setting (\texorpdfstring{$N=1$}{N=1}) with Confuse setup}
We further evaluate our method in the setting where only a single client participates ($N=1$), which reduces to \emph{centralized unlearning}. Our motivation is practical: hyperparameter tuning can be performed efficiently in the single-client (centralized) setting, and the resulting choices can then be broadcast to the full federated setup. This approach reduces computational cost and, in real-world deployments, avoids requiring participation from the entire distributed population of clients during hyperparameter tuning.

We also report additional empirical results for this setting, which provide an expanded treatment of the discussion.

We consider two versions of the model (ResNet-18). The first, which we call the vanilla model, was trained on CIFAR-10, achieving an accuracy of 0.84. We refer to this as the vanilla original model. Notably, this model does not operate at full capacity.
Since we believe that the unlearning frameworks should be independent of the original model’s training procedure, we also evaluate a full-capacity original model, which is a Torchvision pre-trained model with a precision of 0.96. For the unlearning process, we apply one policy. We perform two epochs of maximization, each followed by a minimization step, with an additional minimization step at the end. This is a very extreme recovery treatment with only one round of minimization.

Since $N$ is equal to one we evaluate data level deletion. We consider confuse scenario summarized in Table \ref{tab:forgetting_scenarios}, to support our empirical results. Following setup in \citep{kurmanji2023unboundedmachineunlearning}  which are already mentioned.



\begin{table*}[t]
\centering

\begin{minipage}[t]{0.34\textwidth}
\vspace{0pt}
\centering

\captionof{table}{
Confused-data deletion configurations for the $N=1$ case study.
Each scenario specifies the affected classes and the number of samples
included in the forget set.
}
\label{tab:forgetting_scenarios}

\setlength{\tabcolsep}{5pt}
\renewcommand{\arraystretch}{1.10}

\resizebox{\linewidth}{!}{%
\begin{tabular}{l l c}
\toprule
\textbf{Scenario} & \textbf{Forget set} & \textbf{Samples removed} \\
\midrule
Complete & Entire class 5       & All \\
Moderate & Class 5              & 500 \\
Dual     & Classes 4 and 5      & 500/class \\
Broad    & Classes 1--5         & 500/class \\
Extended & Classes 1--6         & 500/class \\
\bottomrule
\end{tabular}%
}

\end{minipage}
\hfill
\begin{minipage}[t]{0.64\textwidth}
\vspace{0pt}
\centering

\captionof{table}{
CIFAR-10 test error (\%) after data-level unlearning in the $N=1$
centralized setting with ResNet-18.
The Vanilla and Full-Capacity models have pre-unlearning test errors of
16\% and 4\%, respectively. Results are reported as mean $\pm$ standard
deviation; lower is better.
}
\label{tab:forgetting_results_5}

\setlength{\tabcolsep}{5pt}
\renewcommand{\arraystretch}{1.12}

\resizebox{\linewidth}{!}{%
\begin{tabular}{l l cc cc}
\toprule

\multirow{2}{*}{\textbf{Method}}
& \multirow{2}{*}{\textbf{Scenario}}
& \multicolumn{2}{c}{\textbf{Vanilla: Test Error}}
& \multicolumn{2}{c}{\textbf{Full Capacity: Test Error}} \\

\cmidrule(lr){3-4}
\cmidrule(lr){5-6}

&
& \textbf{Pretrain} & \textbf{Unlearn}
& \textbf{Pretrain} & \textbf{Unlearn} \\

\midrule


Ours (KL)
& \multirow{2}{*}{Complete}
& \multirow{2}{*}{16.00}
& 82.28 $\pm$ 7.10
& \multirow{2}{*}{4.00}
& 43.11 $\pm$ 3.99 \\

Ours (JS)
&
&
& 83.01 $\pm$ 2.47
&
& 56.34 $\pm$ 0.69 \\

\cmidrule(lr){2-6}


Ours (KL)
& \multirow{2}{*}{Moderate}
& \multirow{2}{*}{16.00}
& 73.05 $\pm$ 4.55
& \multirow{2}{*}{4.00}
& 18.64 $\pm$ 1.47 \\

Ours (JS)
&
&
& 61.79 $\pm$ 1.90
&
& 21.68 $\pm$ 0.14 \\

\cmidrule(lr){2-6}


Ours (KL)
& \multirow{2}{*}{Dual}
& \multirow{2}{*}{16.00}
& 79.59 $\pm$ 0.21
& \multirow{2}{*}{4.00}
& 26.27 $\pm$ 0.09 \\

Ours (JS)
&
&
& 79.56 $\pm$ 0.09
&
& 34.23 $\pm$ 0.40 \\

\cmidrule(lr){2-6}


Ours (KL)
& \multirow{2}{*}{Broad}
& \multirow{2}{*}{16.00}
& 74.36 $\pm$ 3.71
& \multirow{2}{*}{4.00}
& 59.88 $\pm$ 1.82 \\

Ours (JS)
&
&
& 80.11 $\pm$ 0.07
&
& 59.55 $\pm$ 1.08 \\

\cmidrule(lr){2-6}


Ours (KL)
& \multirow{2}{*}{Extended}
& \multirow{2}{*}{16.00}
& 82.81 $\pm$ 5.48
& \multirow{2}{*}{4.00}
& 55.38 $\pm$ 4.27 \\

Ours (JS)
&
&
& 79.04 $\pm$ 0.47
&
& 63.76 $\pm$ 2.82 \\

\bottomrule
\end{tabular}%
}

\end{minipage}

\end{table*}

We analyze the performance of the models across different confuse scenarios. In this section, we focus on KL-JS and KL-KL method variance. The best performance is achieved when the error in the test dataset is minimized. As shown in Table \ref{tab:forgetting_results_5}, using JS loss as the maximization loss generally results in \textbf{lower variance} across almost most of all scenarios.

To interpret these results, we define the best loss function as the one where test error is lowest. This is not always the case. With a more nuanced analysis, we argue that KL-JS performs better in most scenarios with confidence. In cases where it does not, model degradation complicates the analysis and introduces significant complexity.

In the complete forgetting scenario, where the goal is to forget all data from a specific class, the error on the forget set rapidly reaches 100\%. After two max-min epochs, minimizing with the largest possible batch size yields strong results, as demonstrated in \citep{kurmanji2023unboundedmachineunlearning}. The key point here is that because the entire class is absent, post-minimization does not affect the forget error.

As shown in Table \ref{tab:forgetting_results_5}, in the Moderate case, KL-JS clearly outperforms the baseline in the Vanilla setup. In the full capacity case for Broad, the KL$-$JS outperforming baseline is determined.

Our observations indicate that in the Vanilla case for Moderate (forgetting 500 samples) and up to Extended cases, the degradation in model performance is so severe that one could argue the model has effectively lost its knowledge. This highlights a critical limitation based on current literature 
when the number of deleted samples per class increases. This issue presents emerging challenges in the field, which are highly relevant to real world scenarios. A similar problem is also addressed in \citep{sekhari2021rememberwantforgetalgorithms}. Broad and Extended cases in full capacity also suffer the same problem mentioned here. As you can see, there is not a single case where KL$-$KL outperforms KL$-$JS with full confidence (lower variance) and not degraded model.


\section{Communication-Cost Analysis}
\label{app:communication_cost}

Federated unlearning should be evaluated not only in terms of predictive
utility and forgetting effectiveness, but also in terms of the communication
required to obtain the unlearned model. This distinction is particularly
important because, in federated learning, communication between the server
and clients can dominate the cost of local computation.

We therefore analyze communication using two complementary measures.
The first measure is implementation independent and is the primary measure
used in our experiments. The second provides a more fine-grained analytical
view of the communication generated by \method.

\subsection{Synchronization-Phase Accounting}

A synchronization phase consists of
\[
\text{server broadcast}
\;\rightarrow\;
\text{local client computation}
\;\rightarrow\;
\text{client upload}
\;\rightarrow\;
\text{server aggregation}.
\]

For \method, a single max--min iteration contains two distinct synchronization
phases. The first phase performs forget-set maximization, while the second
performs retained-data minimization. Hence, if $T$ denotes the number of
max--min iterations and $T_{\mathrm{post}}$ denotes the number of
post-training minimization iterations, the total number of synchronization
phases is
\begin{equation}
    B_{\method}
    =
    2T + T_{\mathrm{post}}.
    \label{eq:ffum_comm}
\end{equation}

This accounting is consistent with
Algorithm~\ref{alg:fed-f-scrub-gradient-local}: the server aggregates once
after the maximization update and once again after the retained-data
minimization update.

For comparison, we denote the corresponding phase decomposition of the
baselines as
\begin{align}
    B_{\mathrm{Halimi}}
        &= T_{\mathrm{ascent}} + T_{\mathrm{fine}}, \\
    B_{\mathrm{NoT}}
        &= T_{\mathrm{neg}} + T_{\mathrm{fine}}, \\
    B_{\mathrm{MoDE}}
        &= T_{\mathrm{deg}} + T_{\mathrm{guid}},
    \label{eq:comm_alt}
\end{align}
where $T_{\mathrm{neg}}$ and $T_{\mathrm{fine}}$ refer to the
weight-negation and fine-tuning phases of NoT
\citep{khalil2025notfederatedunlearningweight},
$T_{\mathrm{ascent}}$ and $T_{\mathrm{fine}}$ to the gradient-ascent
and fine-tuning phases of Halimi
\citep{halimi2023federatedunlearningefficientlyerase}, and
$T_{\mathrm{deg}}$ and $T_{\mathrm{guid}}$ to the degradation and
guidance phases of MoDE \citep{10269017}.

To avoid confounding unlearning quality with additional communication, all
approximate-unlearning methods are evaluated under the same predefined
synchronization-phase budget,
\begin{equation}
    B_{\method}
    =
    B_{\mathrm{Halimi}}
    =
    B_{\mathrm{NoT}}
    =
    B_{\mathrm{MoDE}}
    =
    B.
    \label{eq:comm_budget}
\end{equation}

\paragraph{Interpretation of the equal-budget comparison.}
We assign \method and each approximate-unlearning baseline the same
predefined budget $B$ of server--client synchronization phases. A
phase comprises a server broadcast, local client computation, client
upload, and server aggregation. In \method, maximization and
minimization are separate phases, so its total is
\[
2T + T_{\mathrm{post}} = B.
\]

This protocol controls the number of synchronization opportunities
available to each method. Consequently, the head-to-head experiments
compare forgetting effectiveness and retained-model utility at an
equal phase budget; they do not claim that \method uses fewer phases
than the other approximate-unlearning methods. Equal phase counts do
not necessarily imply equal transmitted bytes, local computation, or
wall-clock time. Since these quantities were not measured consistently
for every baseline, we make no head-to-head efficiency claim on those
measures.

Thus, comparisons between approximate-unlearning methods are performed under
equal communication opportunity. The communication advantage relative to
exact retraining arises because exact retraining must repeat the complete
federated training schedule on the retained data. Importantly, we do not
claim a measured transmitted-byte advantage over NoT, Halimi, or MoDE.
Total transmitted bytes were not explicitly logged for these methods and
may depend on implementation details including broadcast mechanisms,
compression, secure aggregation, and client availability. Instead, the
head-to-head experimental comparison controls communication through the
common synchronization budget in Eq.~\eqref{eq:comm_budget}.

The finer-grained analysis in
Eqs.~\eqref{eq:ffum_comm}--\eqref{eq:comm_alt} shows, however, that
synchronization-phase count can be conservative for \method in client-level
unlearning: a forgetting phase may require an upload from only the departing
client, whereas an ordinary retained-data training phase requires updates
from all retained clients.
\section{Theoretical Results}\label{sec:theory-res}
In this section, we provide three main theoretical results for our approach. The first theoretical question is whether \method  is merely a heuristic alternation of ascent and descent steps, or whether it optimizes a well-defined objective.  This point is important because the algorithm maximizes a forget-set divergence and then minimizes a retained-data objective, which could otherwise look like two unrelated procedures.
The second theoretical question is why \method  should alternate forget-set maximization with retained-data minimization.  This is central point because max-only unlearning can move the decision boundary away from the forget set while also destroying the retained-data structure. The third theoretical question is whether the model produced after \method is close to a retained-data optimization target.  Exact unlearning is defined by equivalence to training on $S_R$ \citep{cao2015towards}; in the nonconvex analysis below, we therefore use a local retained-data ERM minimizer as a tractable reference target rather than claiming parameter equality with an arbitrary fresh retraining run. In the following, we first provide notations and preliminaries for our main theoretical results. Then, we provide main theoretical results.
\subsection{Notations}
\label{sec:notation}

We repeat the notations in this section. Let $C$ be the number of clients.  Client $c$ has local data
\begin{equation}
    S^{(c)}=S_R^{(c)}\cup S_F^{(c)},
    \qquad
    S_R^{(c)}\cap S_F^{(c)}=\emptyset,
\end{equation}
where $S_F^{(c)}$ is the local forget set and $S_R^{(c)}$ is the local retained set \footnote{
Throughout, expressions normalized by \(n_R\) or \(n_F\) are understood only when the corresponding set is nonempty. If \(S_R=\emptyset\) or \(S_F=\emptyset\), the associated loss/divergence term is omitted from the objective}.  Define
\begin{equation}
    n_F^{(c)}:=|S_F^{(c)}|,\qquad
    n_R^{(c)}:=|S_R^{(c)}|,
\end{equation}
\begin{equation}
    n_F:=\sum_{c=1}^C n_F^{(c)},
    \qquad
    n_R:=\sum_{c=1}^C n_R^{(c)}.
\end{equation}
Let
\begin{equation}
    q_\theta(x):=h(x;\theta)\in \Delta^{K-1}
\end{equation}
denote the predictive probability vector of the model.  Let $f_F$ be the divergence generator used on the forget set and $f_R$ the generator used on the retained set.  For a teacher or reference parameter $\theta_T$, define
\begin{equation}
    d_F(x;\theta,\theta_T)
    :=D_{f_F}\bigl(q_\theta(x)\Vert q_{\theta_T}(x)\bigr),
    \label{eq:dF}
\end{equation}
\begin{equation}
    d_R(x;\theta,\theta_T)
    :=D_{f_R}\bigl(q_\theta(x)\Vert q_{\theta_T}(x)\bigr).
    \label{eq:dR}
\end{equation}
The client-level forget objective is
\begin{equation}
    \mathcal L^{(c)}_{\max}(\theta;\theta_T)
    :=
    \frac{1}{n_F^{(c)}}
    \sum_{x\in S_F^{(c)}} d_F(x;\theta,\theta_T),
    \label{eq:client-forget}
\end{equation}
and the client-level retain objective is
\begin{equation}
    \mathcal L^{(c)}_{\min}(\theta;\theta_T)
    :=
    \frac{\alpha}{n_R^{(c)}}
    \sum_{x\in S_R^{(c)}}d_R(x;\theta,\theta_T)
    +
    \frac{\gamma}{n_R^{(c)}}
    \sum_{(x,y)\in S_R^{(c)}}\ell_{f_R}(q_\theta(x),y).
    \label{eq:client-retain}
\end{equation}
The corresponding global objectives are
\begin{equation}
    \mathcal L_{\max}(\theta;\theta_T)
    :=
    \sum_{c=1}^C \frac{n_F^{(c)}}{n_F}\mathcal L^{(c)}_{\max}(\theta;\theta_T)
    =
    \frac{1}{n_F}\sum_{c=1}^C\sum_{x\in S_F^{(c)}}d_F(x;\theta,\theta_T),
    \label{eq:global-forget}
\end{equation}
\begin{equation}
    \mathcal L_{\min}(\theta;\theta_T)
    :=
    \sum_{c=1}^C \frac{n_R^{(c)}}{n_R}\mathcal L^{(c)}_{\min}(\theta;\theta_T)
    \label{eq:global-retain-def}
\end{equation}
with the explicit form
\begin{equation}
    \mathcal L_{\min}(\theta;\theta_T)
    =
    \frac{\alpha}{n_R}\sum_{c=1}^C\sum_{x\in S_R^{(c)}}d_R(x;\theta,\theta_T)
    +
    \frac{\gamma}{n_R}\sum_{c=1}^C\sum_{(x,y)\in S_R^{(c)}}\ell_{f_R}(q_\theta(x),y).
    \label{eq:global-retain}
\end{equation}
Define the contrastive unlearning potential
\begin{equation}
    \Psi_\rho(\theta;\theta_T)
    :=
    \mathcal L_{\min}(\theta;\theta_T)-\rho\mathcal L_{\max}(\theta;\theta_T),
    \qquad
    \rho:=\frac{\eta_{\max}}{\eta_{\min}}.
    \label{eq:contrastive-potential}
\end{equation}
The retain term preserves utility and output behavior on $S_R$, while the negative forget term encourages the unlearned model to move away from the original model on $S_F$.

\subsection{Preliminaries}
\label{subsec:preliminaries-main-results}

This subsection collects the definitions, assumptions, and auxiliary lemmas used by the three main results.  The preliminaries are grouped according to the result they support.

\subsubsection{Federated contrastive optimization}
\label{subsubsec:prelim-result-one}

The contrastive unlearning potential is
\begin{equation}
    \Psi_\rho(\theta;\theta_T)
    :=
    \calR(\theta;\theta_T)-\rho\calF(\theta;\theta_T),
    \qquad
    \rho=\frac{\eta_{\max}}{\eta_{\min}}.
    \label{eq:contrastive-potential-revised}
\end{equation}
The term $\calR$ preserves retained-data behavior and supervised utility, while $-\rho\calF$ creates a directed removal pressure on the forget set.  The federated averaging notation follows FedAvg \citep{mcmahan2017communication}; the retain/forget split  adapts standard machine-unlearning and federated-unlearning formulations 
\citep{cao2015towards,sekhari2021rememberwantforgetalgorithms,romandini2024federated}.

\begin{assumption}[Smoothness]
\label{ass:smoothness-revised}
The functions $\calF$ and $\calR$ are differentiable and have Lipschitz gradients on the local parameter region visited by \method .  That is, there exist $L_F,L_R>0$ such that
\begin{equation}
    \|\nabla \calF(\theta)-\nabla \calF(\theta')\|
    \leq
    L_F\|\theta-\theta'\|,
\end{equation}
\begin{equation}
    \|\nabla \calR(\theta)-\nabla \calR(\theta')\|
    \leq
    L_R\|\theta-\theta'\|.
\end{equation}
Consequently, $\Psi_\rho$ is $L_\Psi$-smooth with
\begin{equation}
    L_\Psi\leq L_R+\rho L_F.
\end{equation}
\end{assumption}

\begin{assumption}[Bounded gradients]
\label{ass:bounded-gradients-revised}
There exists $B_F>0$ such that
\begin{equation}
    \|\nabla\calF(\theta)\|\leq B_F
\end{equation}
for all iterates considered in the analysis.
\end{assumption}

\begin{lemma}[FedAvg equivalence for one full-batch \method  step]
\label{lem:fedavg-equivalence-revised}
Assume each client performs one full-batch forget-ascent step
\begin{equation}
    \theta_{c,+}^{F}
    =
    \theta_t+\eta_{\max}\nabla\calF^c(\theta_t;\theta_T).
    \label{eq:local-forget-step-revised}
\end{equation}
If the server aggregates the resulting client models with weights $n_F^{(c)}/n_F$, then
\begin{equation}
    \theta_{t+1/2}
    =
    \sum_{c=1}^C\frac{n_F^{(c)}}{n_F}\theta_{c,+}^{F}
    =
    \theta_t+\eta_{\max}\nabla\calF(\theta_t;\theta_T).
    \label{eq:global-forget-step-revised}
\end{equation}
Similarly, if each client performs one full-batch retain-descent step
\begin{equation}
    \theta_{c,+}^{R}
    =
    \theta_{t+1/2}-\eta_{\min}\nabla\calR^c(\theta_{t+1/2};\theta_T)
    \label{eq:local-retain-step-revised}
\end{equation}
and the server aggregates with weights $n_R^{(c)}/n_R$, then
\begin{equation}
    \theta_{t+1}
    =
    \sum_{c=1}^C\frac{n_R^{(c)}}{n_R}\theta_{c,+}^{R}
    =
    \theta_{t+1/2}-\eta_{\min}\nabla\calR(\theta_{t+1/2};\theta_T).
    \label{eq:global-retain-step-revised}
\end{equation}
\end{lemma}

\noindent
This lemma is an exact identity only for one common-start, full-batch local step.  With multiple local steps, client iterates generally drift apart and the aggregated update acquires an additional client-drift error.

\begin{proof}
By the definition of $\calF$,
\begin{align}
    \sum_{c=1}^C\frac{n_F^{(c)}}{n_F}\theta_{c,+}^{F}
    &=
    \sum_{c=1}^C\frac{n_F^{(c)}}{n_F}
    \left(\theta_t+\eta_{\max}\nabla\calF^c(\theta_t;\theta_T)\right) \\
    &=
    \theta_t+
    \eta_{\max}\sum_{c=1}^C\frac{n_F^{(c)}}{n_F}\nabla\calF^c(\theta_t;\theta_T) \\
    &=
    \theta_t+\eta_{\max}\nabla\calF(\theta_t;\theta_T).
\end{align}
The retained-data step follows identically from the definition of $\calR$.
\end{proof}

\subsubsection{Local \texorpdfstring{$f$}{f}-divergence geometry}
\label{subsubsec:prelim-result-two}

 Let
\begin{equation}
    J_x:=\nabla_\theta q_\theta(x)\big|_{\theta=\theta_T}\in\R^{K\times d}
\end{equation}
be the Jacobian of the model output at the teacher parameter.  Define the empirical Fisher-type matrices
\begin{equation}
    I_R(\theta_T)
    :=
    \frac{1}{n_R}\sum_{x\in S_R}
    J_x^\top\diag(q_{\theta_T}(x))^{-1}J_x,
    \label{eq:IR-revised}
\end{equation}
\begin{equation}
    I_F(\theta_T)
    :=
    \frac{1}{n_F}\sum_{x\in S_F}
    J_x^\top\diag(q_{\theta_T}(x))^{-1}J_x.
    \label{eq:IF-revised}
\end{equation}
Finally, let
\begin{equation}
    H_R:=\nabla^2\calL_R(\theta_T),
    \qquad
    \calL_R(\theta):=\frac{1}{n_R}\sum_{(x,y)\in S_R}\ell(q_\theta(x),y),
    \label{eq:HR-revised}
\end{equation}
and let
\begin{equation}
    c_R:=f_R''(1),
    \qquad
    c_F:=f_F''(1).
\end{equation}

\begin{assumption}[Interior softmax probabilities]
\label{ass:interior-revised}
There exists $\kappa>0$ such that, for all relevant samples $x$, classes $k$, and parameters $\theta$ in the local region of interest,
\begin{equation}
    q_{\theta,k}(x)\geq \kappa.
\end{equation}
This local condition is automatically satisfied if probabilities are clipped away from zero.
\end{assumption}

\begin{assumption}[Local differentiability of the predictive map]
\label{ass:local-C2-revised}
For every relevant sample $x$, the map $\theta\mapsto q_\theta(x)$ is twice continuously differentiable on a neighborhood of $\theta_T$.  More generally, it is sufficient that the iterates remain in a local region on which the first- and second-order derivatives used below exist and are continuous.
\end{assumption}

\begin{lemma}[Local Fisher expansion of $f$-divergences \citep{polyanskiy2014lecture}]
\label{lem:fisher-expansion-revised}
Assume Assumptions~\ref{ass:interior-revised} and~\ref{ass:local-C2-revised}.  Let $f$ be twice differentiable at $1$, with $f(1)=0$, and normalize it so that $f'(1)=0$.  For a small perturbation $\delta\in\R^d$,
\begin{equation}
    D_f\bigl(q_{\theta_T+\delta}(x)\Vert q_{\theta_T}(x)\bigr)
    =
    \frac{f''(1)}{2}
    \delta^\top
    J_x^\top\diag(q_{\theta_T}(x))^{-1}J_x
    \delta
    +o(\|\delta\|^2).
    \label{eq:fisher-expansion-revised}
\end{equation}
For the divergences used in \method ,
\begin{equation}
    f_{\KL}''(1)=1,
    \qquad
    f_{\JS}''(1)=\frac{1}{4},
    \qquad
    f_{\chi^2}''(1)=2.
    \label{eq:fpp-values-revised}
\end{equation}
Moreover, with natural logarithms,
\begin{equation}
    0\leq D_{\JS}(p\Vert q)\leq \log 2.
    \label{eq:js-bounded-revised}
\end{equation}
\end{lemma}


\subsubsection{Retraining target}
\label{subsubsec:prelim-result-three}

We use the retained-data empirical-risk minimization (ERM) target
\begin{equation}
    \theta_R^\star:=\argmin_\theta \calL_R(\theta),
    \label{eq:theta-R-star-revised}
\end{equation}
where $\calL_R$ is the retained supervised loss in Equation~\eqref{eq:HR-revised}.  This target is a local optimization reference for the analysis; in a nonconvex neural network it need not coincide in parameter space with the output of rerunning the full stochastic federated training algorithm on $S_R$.  The post-training objective is
\begin{equation}
    \calP_\alpha(\theta)
    :=
    \gamma\calL_R(\theta)+\alpha\calD_R(\theta;\theta_T),
    \label{eq:post-objective-revised}
\end{equation}
where
\begin{equation}
    \calD_R(\theta;\theta_T)
    :=
    \frac{1}{n_R}\sum_{x\in S_R}
    D_{f_R}\bigl(q_\theta(x)\Vert q_{\theta_T}(x)\bigr).
    \label{eq:DR-revised}
\end{equation}
Let
\begin{equation}
    \theta_\alpha^\star:=\argmin_\theta \calP_\alpha(\theta).
    \label{eq:theta-alpha-star-revised}
\end{equation}
The following local optimization assumptions are standard for analyzing optimization problems.

\begin{assumption}[Local strong convexity for distance guarantees]
\label{ass:strong-convexity-revised}
There exists a convex local neighborhood $\mathcal B$ containing the relevant post-training iterates and the local minimizers $\theta_\alpha^\star$ and $\theta_R^\star$ such that $\calP_\alpha$ is $\mu_P$-strongly convex on $\mathcal B$.  In particular, whenever Proposition~\ref{thm:main-post-training-revised} is invoked, $\theta_U$, $\theta_\alpha^\star$, and $\theta_R^\star$, together with the line segments joining them, are assumed to lie in $\mathcal B$.  In the special case $\alpha=0$, $\calL_R$ is locally $\mu_R$-strongly convex on the same neighborhood, so that $\gamma\calL_R$ is $\gamma\mu_R$-strongly convex.
\end{assumption}

\begin{assumption}[Local PL condition for optimization-rate guarantees]
\label{ass:pl-revised}
The post-training objective $\calP_\alpha$ is $L_P$-smooth and satisfies the Polyak--Lojasiewicz condition in the local basin reached by \method : for some $\mu_{PL}>0$,
\begin{equation}
    \frac{1}{2}\|\nabla\calP_\alpha(\theta)\|^2
    \geq
    \mu_{PL}\bigl(\calP_\alpha(\theta)-\calP_\alpha^\star\bigr).
    \label{eq:PL-revised}
\end{equation}
\end{assumption}

\begin{lemma}[Evaluation transfer from parameter/output closeness]
\label{lem:evaluation-transfer-revised}
The following two consequences will be used to interpret approximate retraining.

\begin{enumerate}[label=(\alph*),leftmargin=*]
    \item If the supervised loss is cross-entropy,
    \begin{equation}
        \ell(q_\theta(x),y)=-\log q_{\theta,y}(x),
    \end{equation}
    then the empirical retained-data classification error
    \begin{equation}
        \widehat{\mathrm{err}}_R(\theta)
        :=
        \frac{1}{n_R}\sum_{(x,y)\in S_R}
        \one\left[\argmax_k q_{\theta,k}(x)\neq y\right]
    \end{equation}
    satisfies
    \begin{equation}
        \widehat{\mathrm{err}}_R(\theta)
        \leq
        \frac{\calL_R(\theta)}{\log 2}.
        \label{eq:ce-error-transfer-revised}
    \end{equation}

    \item Let $a:\Delta^{K-1}\to[0,1]$ be an output-based membership-inference attack score that is $L_a$-Lipschitz in $\ell_1$ distance:
    \begin{equation}
        |a(p)-a(q)|\leq L_a\|p-q\|_1.
        \label{eq:attack-Lipschitz-revised}
    \end{equation}
    Assume also that the model output is locally $L_h$-Lipschitz in parameters:
    \begin{equation}
        \|q_\theta(x)-q_{\theta'}(x)\|_1
        \leq
        L_h\|\theta-\theta'\|.
        \label{eq:model-Lipschitz-revised}
    \end{equation}
    Then, for any forget set $S_F$,
    \begin{align}
        &\left|
        \frac{1}{n_F}\sum_{x\in S_F}a(q_{\theta}(x))
        -
        \frac{1}{n_F}\sum_{x\in S_F}a(q_{\theta'}(x))
        \right|
        \nonumber\\
        &\qquad\leq
        L_aL_h\|\theta-\theta'\|.
        \label{eq:mia-transfer-revised}
    \end{align}
\end{enumerate}
\end{lemma}

\begin{proof}
For part (a), if the classifier predicts incorrectly, then the true-class probability cannot exceed $1/2$, i.e., $q_{\theta,y}(x)\leq 1/2$.  Hence $-\log q_{\theta,y}(x)\geq \log 2$, and
\begin{equation}
    \one\left[\argmax_k q_{\theta,k}(x)\neq y\right]
    \leq
    \frac{-\log q_{\theta,y}(x)}{\log 2}.
\end{equation}
Averaging over $S_R$ proves Equation~\eqref{eq:ce-error-transfer-revised}.

For part (b), by the triangle inequality, the Lipschitz property of $a$, and the Lipschitz property of $q_\theta$,
\begin{align}
    &\left|
    \frac{1}{n_F}\sum_{x\in S_F}a(q_{\theta}(x))
    -
    \frac{1}{n_F}\sum_{x\in S_F}a(q_{\theta'}(x))
    \right| \\
    &\qquad\leq
    \frac{1}{n_F}\sum_{x\in S_F}
    |a(q_\theta(x))-a(q_{\theta'}(x))| \\
    &\qquad\leq
    \frac{L_a}{n_F}\sum_{x\in S_F}
    \|q_\theta(x)-q_{\theta'}(x)\|_1 \\
    &\qquad\leq
    L_aL_h\|\theta-\theta'\|.
\end{align}
\end{proof}

The classification-error bound is an instance of the standard implication from surrogate-risk consistency to classification-risk consistency; a general treatment of surrogate calibration is given by \citep{bartlett2006convexity}.  The membership-inference part is stated for output-based attacks, following the attack setting introduced by \citep{hu2022membership}.  The need to evaluate unlearning against the retained-data retraining model is emphasized by \citep{hayes2024inexactunlearningneedscareful}.

\subsection{Main Results}
\subsubsection{ \texorpdfstring{$\method$}{method} is an inexact optimizer of a contrastive unlearning objective}
\label{subsec:main-result-one}

The result below shows that, after the weighted aggregation justified in Lemma~\ref{lem:fedavg-equivalence-revised}, one idealized \method round is an inexact gradient step on the contrastive potential $\Psi_\rho=\calR-\rho\calF$.  In the unlearning context, this result is needed to justify that \method has a principled optimization target rather than only empirical motivation.

Throughout the theoretical analysis in this subsection, the teacher parameter is fixed across unlearning rounds.  In particular, one may take
\begin{equation}
    \theta_T=\theta_0.
    \label{eq:fixed-teacher-theory}
\end{equation}
A round-dependent teacher produces a time-varying objective and requires an additional teacher-drift term; that setting is not covered by the stationarity result below.

\begin{theorem}[Inexact contrastive optimization for the fixed-teacher one-step update]
\label{thm:main-inexact-stationarity-revised}
Let $\eta:=\eta_{\min}$ and $\rho:=\eta_{\max}/\eta_{\min}$.  Consider the idealized aggregated \method  round
\begin{equation}
    \theta_{t+1/2}
    =
    \theta_t+\eta_{\max}\nabla\calF(\theta_t;\theta_T),
    \label{eq:ideal-half-main-revised}
\end{equation}
\begin{equation}
    \theta_{t+1}
    =
    \theta_{t+1/2}-\eta_{\min}\nabla\calR(\theta_{t+1/2};\theta_T).
    \label{eq:ideal-full-main-revised}
\end{equation}
Under Assumptions~\ref{ass:smoothness-revised} and~\ref{ass:bounded-gradients-revised}, the update can be written as
\begin{equation}
    \theta_{t+1}
    =
    \theta_t-\eta\nabla\Psi_\rho(\theta_t;\theta_T)+e_t,
    \label{eq:inexact-main-revised}
\end{equation}
where
\begin{equation}
    \|e_t\|\leq \rho\eta^2 L_RB_F.
    \label{eq:error-main-revised}
\end{equation}
Furthermore, suppose that the fixed-teacher potential $\Psi_\rho(\cdot;\theta_T)$ is lower bounded by $\Psi_{\inf}$ and that the inexact update satisfies the more general bound $\|e_t\|\leq \eta\varepsilon_{\mathrm{alg}}$.  If
\begin{equation}
    0<\eta\leq \frac{1}{2L_\Psi},
\end{equation}
then after $T$ rounds,
\begin{equation}
    \frac{1}{T}\sum_{t=0}^{T-1}\|\nabla_\theta\Psi_\rho(\theta_t;\theta_T)\|^2
    \leq
    \frac{4(\Psi_\rho(\theta_0;\theta_T)-\Psi_{\inf})}{\eta T}
    +8\varepsilon_{\mathrm{alg}}^2,
    \label{eq:stationarity-main-revised}
\end{equation}
and therefore
\begin{equation}
    \min_{0\leq t<T}\|\nabla_\theta\Psi_\rho(\theta_t;\theta_T)\|^2
    \leq
    \frac{4(\Psi_\rho(\theta_0;\theta_T)-\Psi_{\inf})}{\eta T}
    +8\varepsilon_{\mathrm{alg}}^2.
    \label{eq:min-stationarity-main-revised}
\end{equation}
For the one-step full-batch version in Equations~\eqref{eq:ideal-half-main-revised}--\eqref{eq:ideal-full-main-revised}, one may take
\begin{equation}
    \varepsilon_{\mathrm{alg}}=\rho\eta L_RB_F.
\end{equation}
\end{theorem}

\begin{proof}
The analysis is inspired by standard smooth nonconvex optimization tools \citep{ghadimi2013stochasticfirstzerothordermethods,bottou2018optimizationmethodslargescalemachine}. 

We first prove the inexact-gradient representation.  From Equations~\eqref{eq:ideal-half-main-revised}--\eqref{eq:ideal-full-main-revised},
\begin{equation}
    \theta_{t+1}
    =
    \theta_t+
    \rho\eta\nabla\calF(\theta_t;\theta_T)
    -
    \eta\nabla\calR(\theta_{t+1/2};\theta_T).
\end{equation}
Adding and subtracting $\eta\nabla\calR(\theta_t;\theta_T)$ gives
\begin{equation}
    \theta_{t+1}
    =
    \theta_t-
    \eta\left(\nabla\calR(\theta_t;\theta_T)-\rho\nabla\calF(\theta_t;\theta_T)\right)
    +e_t,
\end{equation}
where
\begin{equation}
    e_t:=
    -\eta\left(\nabla\calR(\theta_{t+1/2};\theta_T)-\nabla\calR(\theta_t;\theta_T)\right).
\end{equation}
Since $\nabla\Psi_\rho=\nabla\calR-\rho\nabla\calF$, Equation~\eqref{eq:inexact-main-revised} follows.  By $L_R$-smoothness of $\calR$,
\begin{align}
    \|e_t\|
    &\leq
    \eta L_R\|\theta_{t+1/2}-\theta_t\| \\
    &=
    \eta L_R\eta_{\max}\|\nabla\calF(\theta_t;\theta_T)\| \\
    &\leq
    \rho\eta^2 L_RB_F,
\end{align}
which proves Equation~\eqref{eq:error-main-revised}.

We next prove the stationarity bound.  Let $g_t:=\nabla_\theta\Psi_\rho(\theta_t;\theta_T)$.  By $L_\Psi$-smoothness of the fixed-teacher potential,
\begin{equation}
    \Psi_\rho(\theta_{t+1};\theta_T)
    \leq
    \Psi_\rho(\theta_t;\theta_T)
    +\langle g_t,\theta_{t+1}-\theta_t\rangle
    +\frac{L_\Psi}{2}\|\theta_{t+1}-\theta_t\|^2.
\end{equation}
Using $\theta_{t+1}-\theta_t=-\eta g_t+e_t$ and $\|-\eta g_t+e_t\|^2\leq 2\eta^2\|g_t\|^2+2\|e_t\|^2$, we obtain
\begin{align}
    \Psi_\rho(\theta_{t+1};\theta_T)
    &\leq
    \Psi_\rho(\theta_t;\theta_T)
    -\eta\|g_t\|^2
    +\langle g_t,e_t\rangle
    +L_\Psi\eta^2\|g_t\|^2
    +L_\Psi\|e_t\|^2.
\end{align}
Because $\eta\leq 1/(2L_\Psi)$,
\begin{equation}
    -\eta\|g_t\|^2+L_\Psi\eta^2\|g_t\|^2
    \leq
    -\frac{\eta}{2}\|g_t\|^2.
\end{equation}
Using $\|e_t\|\leq\eta\varepsilon_{\mathrm{alg}}$ and Young's inequality,
\begin{equation}
    \langle g_t,e_t\rangle
    \leq
    \eta\varepsilon_{\mathrm{alg}}\|g_t\|
    \leq
    \frac{\eta}{4}\|g_t\|^2+
    \eta\varepsilon_{\mathrm{alg}}^2.
\end{equation}
Also,
\begin{equation}
    L_\Psi\|e_t\|^2
    \leq
    L_\Psi\eta^2\varepsilon_{\mathrm{alg}}^2
    \leq
    \frac{\eta}{2}\varepsilon_{\mathrm{alg}}^2.
\end{equation}
Therefore,
\begin{equation}
    \Psi_\rho(\theta_{t+1};\theta_T)
    \leq
    \Psi_\rho(\theta_t;\theta_T)
    -\frac{\eta}{4}\|g_t\|^2
    +2\eta\varepsilon_{\mathrm{alg}}^2.
\end{equation}
Summing from $t=0$ to $T-1$ and using the lower bound
$\Psi_\rho(\theta^{(T)};\theta_T)\geq\Psi_{\inf}$ for the final iterate
$\theta^{(T)}$ gives
\begin{equation}
    \frac{\eta}{4}\sum_{t=0}^{T-1}\|g_t\|^2
    \leq
    \Psi_\rho(\theta_0;\theta_T)-\Psi_{\inf}
    +2\eta T\varepsilon_{\mathrm{alg}}^2.
\end{equation}
Dividing by $\eta T/4$ proves Equation~\eqref{eq:stationarity-main-revised}.  Equation~\eqref{eq:min-stationarity-main-revised} follows because the minimum is no larger than the average.
\end{proof}

Note that the bound in Theorem~\ref{thm:main-inexact-stationarity-revised} is for the idealized one-full-batch-step update.  With mini-batches or multiple local epochs, additional stochastic-gradient and client-drift terms enter the inexactness error.  Quantifying those terms requires further variance and client-heterogeneity assumptions and is outside the present theorem.

\begin{remark}[Lower boundedness for JS forget maximization]
\label{rem:js-lower-bound-revised}
When the forget objective uses the standard Jensen--Shannon divergence, $0\leq D_{\JS}\leq\log 2$, and hence $\calF(\theta;\theta_T)\leq\log 2$.  Since the retained-data objective $\calR$ is composed of the nonnegative divergence and supervised-loss terms used here, $\calR(\theta;\theta_T)\geq 0$,
\begin{equation}
    \Psi_\rho(\theta;\theta_T)
    =
    \calR(\theta;\theta_T)-\rho\calF(\theta;\theta_T)
    \geq
    -\rho\log 2.
\end{equation}
Thus, for JS forget maximization the lower-boundedness assumption in Theorem~\ref{thm:main-inexact-stationarity-revised} holds automatically with $\Psi_{\inf}\geq-\rho\log 2$.  For unbounded forget divergences such as KL or $\chi^2$, lower boundedness must instead be imposed through the local region of analysis or an appropriate bounded-output/parameter assumption.
\end{remark}

\subsubsection{Retained-data minimization prevents local collapse}
\label{subsec:main-result-two}

 The toy example illustrates this phenomenon visually: the max-only phase can collapse the clean decision boundary, whereas the min--max phase preserves the global class structure and leaves only a local distortion to be repaired by post-training. The following Theorem shows that forget-set maximization introduces negative curvature in forget-sensitive directions, while the retained-data objective contributes positive curvature that can dominate this destabilizing effect.

\begin{theorem}[Local curvature stabilization by retained-data minimization]
\label{thm:main-retention-stability-revised}
Under Assumptions~\ref{ass:interior-revised} and~\ref{ass:local-C2-revised} and the conditions of Lemma~\ref{lem:fisher-expansion-revised}, the local Hessian of the contrastive objective around $\theta_T$ is
\begin{equation}
    \nabla^2\Psi_\rho(\theta_T)
    =
    \gamma H_R+
    \alpha c_R I_R(\theta_T)-
    \rho c_F I_F(\theta_T).
    \label{eq:contrastive-hessian-main-revised}
\end{equation}
If
\begin{equation}
    \lambda_{\min}\left(\gamma H_R+\alpha c_RI_R(\theta_T)\right)
    >
    \rho c_F\lambda_{\max}\left(I_F(\theta_T)\right),
    \label{eq:stability-condition-main-revised}
\end{equation}
where $\lambda_{\min}(\cdot)$ and $\lambda_{\max}(\cdot)$ are minimum eigenvalue and maximum eigenvalue, respectively.
then
\begin{equation}
    \nabla^2\Psi_\rho(\theta_T)\succ 0.
\end{equation}
Equivalently, positive local curvature is guaranteed whenever
\begin{equation}
    \rho
    <
    \frac{\lambda_{\min}\left(\gamma H_R+\alpha c_RI_R(\theta_T)\right)}
    {c_F\lambda_{\max}\left(I_F(\theta_T)\right)}.
    \label{eq:rho-stability-main-revised}
\end{equation}
If the retained-data objective is removed, i.e., $\alpha=\gamma=0$, then
\begin{equation}
    \nabla^2\Psi_\rho(\theta_T)
    =
    -\rho c_FI_F(\theta_T),
\end{equation}
which is negative semidefinite and has strictly negative curvature along any forget-sensitive direction $v$ satisfying $v^\top I_F(\theta_T)v>0$.  Thus, pure divergence maximization contributes destabilizing curvature in forget-sensitive directions, whereas the retained-data objective acts as a curvature stabilizer.
\end{theorem}

\begin{proposition}[Restricted curvature version]
\label{rem:restricted-curvature-revised}
The full-space condition in Equation~\eqref{eq:stability-condition-main-revised} is sufficient and may be conservative in heavily overparameterized networks, where the minimum eigenvalue of the full Hessian can be zero.  The same argument applies on any subspace with orthonormal basis $U$.  In particular, positive curvature on that subspace is guaranteed if
\begin{equation}
\lambda_{\min}\!\left(
U^\top\!\left[\gamma H_R+\alpha c_R I_R(\theta_T)\right]U
\right)
>
\rho c_F\,
\lambda_{\max}\!\left(U^\top I_F(\theta_T)U\right).
\end{equation}
A natural choice is a forget-sensitive subspace contained in the range of $I_F(\theta_T)$.
\end{proposition}

\begin{proof}
By Lemma~\ref{lem:fisher-expansion-revised}, the Hessian of the forget divergence at $\theta_T$ is $c_FI_F(\theta_T)$, and the Hessian of the retained divergence at $\theta_T$ is $c_RI_R(\theta_T)$.  The supervised retained loss contributes $H_R$.  Since
\begin{equation}
    \Psi_\rho=\calR-\rho\calF,
\end{equation}
the Hessian is exactly Equation~\eqref{eq:contrastive-hessian-main-revised}.

For any unit vector $v$,
\begin{align}
    v^\top\nabla^2\Psi_\rho(\theta_T)v
    &=
    v^\top\left(\gamma H_R+\alpha c_RI_R(\theta_T)\right)v
    -
    \rho c_F v^\top I_F(\theta_T)v \\
    &\geq
    \lambda_{\min}\left(\gamma H_R+\alpha c_RI_R(\theta_T)\right)
    -
    \rho c_F\lambda_{\max}\left(I_F(\theta_T)\right).
\end{align}
If Equation~\eqref{eq:stability-condition-main-revised} holds, the right-hand side is positive for every unit vector $v$, which proves positive definiteness.  
\end{proof}

\begin{remark}[Why JS often gives more stable maximization]
\label{rem:js-stability-revised}
Equation~\eqref{eq:rho-stability-main-revised} shows that the destabilizing curvature scales with $c_F=f_F''(1)$.  Since
\begin{equation}
    c_F=\frac{1}{4}\quad\text{for JS},
    \qquad
    c_F=1\quad\text{for KL},
    \qquad
    c_F=2\quad\text{for }\chi^2,
\end{equation}
JS contributes less local negative curvature than KL and $\chi^2$.  In addition, $D_{\JS}$ is bounded by $\log 2$.  Therefore, JS can be interpreted as a more conservative forget-set maximization loss, which supports the empirical observation that KL-JS variants can be more stable in difficult unlearning settings.  This is a local curvature interpretation, not a claim that JS is uniformly optimal in all regimes.
\end{remark}

\subsubsection{Post-training approaches the retained-data ERM target}
\label{subsec:main-result-three}

 The alternating min-max phase is designed to remove the forget-set influence while avoiding collapse, but it may still leave residual distortion.  The post-training phase is therefore needed to move the model back toward the retained-data solution.  The result below formalizes this role: under a local strong-convexity condition, the final distance to the retained-data ERM target $\theta_R^\star$ is controlled by an optimization error term and a bias term caused by retaining the teacher-divergence penalty.  The PL corollary then justifies the empirical use of only a small number of post-training rounds. 

\begin{proposition}[Local closeness to the retained-data ERM target after post-training]
\label{thm:main-post-training-revised}
Let $\theta_R^\star$ and $\theta_\alpha^\star$ be defined by Equations~\eqref{eq:theta-R-star-revised} and~\eqref{eq:theta-alpha-star-revised}.  Assume Assumption~\ref{ass:strong-convexity-revised}.  Suppose also that
\begin{equation}
    \|\nabla\calD_R(\theta_R^\star;\theta_T)\|\leq B_D.
    \label{eq:BD-main-revised}
\end{equation}
If post-training returns $\theta_U$ satisfying
\begin{equation}
    \calP_\alpha(\theta_U)-\calP_\alpha(\theta_\alpha^\star)
    \leq
    \varepsilon_{\mathrm{min}},
    \label{eq:post-error-main-revised}
\end{equation}
then
\begin{equation}
    \|\theta_U-\theta_R^\star\|
    \leq
    \sqrt{\frac{2\varepsilon_{\mathrm{min}}}{\mu_P}}
    +
    \frac{\alpha B_D}{\mu_P}.
    \label{eq:post-distance-main-revised}
\end{equation}
If the final post-training objective sets $\alpha=0$ and returns $\theta_U$ satisfying
\begin{equation}
    \calP_0(\theta_U)-\calP_0(\theta_R^\star)
    \leq
    \varepsilon_{\mathrm{min}},
    \qquad
    \calP_0=\gamma\calL_R,
    \label{eq:post-error-alpha-zero-revised}
\end{equation}
then the teacher-divergence bias term is absent and
\begin{equation}
    \|\theta_U-\theta_R^\star\|
    \leq
    \sqrt{\frac{2\varepsilon_{\mathrm{min}}}{\gamma\mu_R}}.
    \label{eq:post-distance-alpha-zero-revised}
\end{equation}
Furthermore, under the Lipschitz conditions in Lemma~\ref{lem:evaluation-transfer-revised}, the difference between any Lipschitz output-based membership score evaluated at $\theta_U$ and at the retained-data ERM target $\theta_R^\star$ is bounded by
\begin{align}
    &\left|
    \frac{1}{n_F}\sum_{x\in S_F}a(q_{\theta_U}(x))
    -
    \frac{1}{n_F}\sum_{x\in S_F}a(q_{\theta_R^\star}(x))
    \right|
    \nonumber\\
    &\qquad\leq
    L_aL_h
    \left(
    \sqrt{\frac{2\varepsilon_{\mathrm{min}}}{\mu_P}}
    +
    \frac{\alpha B_D}{\mu_P}
    \right).
    \label{eq:mia-main-revised}
\end{align}
Thus, post-training gives a local parameter-space closeness guarantee to the retained-data ERM target, and this closeness transfers to Lipschitz output-based membership scores.  This statement does not by itself cover discontinuous threshold attacks or establish equality with the parameter vector returned by a fresh nonconvex federated retraining run.
\end{proposition}

\begin{proof}
By $\mu_P$-strong convexity of $\calP_\alpha$,
\begin{equation}
    \calP_\alpha(\theta_U)-\calP_\alpha(\theta_\alpha^\star)
    \geq
    \frac{\mu_P}{2}\|\theta_U-\theta_\alpha^\star\|^2.
\end{equation}
Combining this inequality with Equation~\eqref{eq:post-error-main-revised} gives
\begin{equation}
    \|\theta_U-\theta_\alpha^\star\|
    \leq
    \sqrt{\frac{2\varepsilon_{\mathrm{min}}}{\mu_P}}.
    \label{eq:post-to-alpha-main-revised}
\end{equation}
Next, because $\theta_R^\star$ minimizes $\calL_R$ locally,
\begin{equation}
    \nabla\calL_R(\theta_R^\star)=0.
\end{equation}
Therefore,
\begin{equation}
    \nabla\calP_\alpha(\theta_R^\star)
    =
    \alpha\nabla\calD_R(\theta_R^\star;\theta_T).
\end{equation}
Since $\theta_\alpha^\star$ minimizes $\calP_\alpha$,
\begin{equation}
    \nabla\calP_\alpha(\theta_\alpha^\star)=0.
\end{equation}
Strong monotonicity of the gradient of a $\mu_P$-strongly convex differentiable function gives
\begin{equation}
    \mu_P\|\theta_\alpha^\star-\theta_R^\star\|
    \leq
    \|\nabla\calP_\alpha(\theta_R^\star)-\nabla\calP_\alpha(\theta_\alpha^\star)\|.
\end{equation}
Hence,
\begin{equation}
    \|\theta_\alpha^\star-\theta_R^\star\|
    \leq
    \frac{\alpha\|\nabla\calD_R(\theta_R^\star;\theta_T)\|}{\mu_P}
    \leq
    \frac{\alpha B_D}{\mu_P}.
    \label{eq:alpha-bias-main-revised}
\end{equation}
Combining Equations~\eqref{eq:post-to-alpha-main-revised} and~\eqref{eq:alpha-bias-main-revised} using the triangle inequality proves Equation~\eqref{eq:post-distance-main-revised}.  For the special case $\alpha=0$, Equation~\eqref{eq:post-error-alpha-zero-revised} and $\gamma\mu_R$-strong convexity of $\calP_0=\gamma\calL_R$ directly give Equation~\eqref{eq:post-distance-alpha-zero-revised}.  Finally, substituting Equation~\eqref{eq:post-distance-main-revised} into the MIA transfer bound in Lemma~\ref{lem:evaluation-transfer-revised} proves Equation~\eqref{eq:mia-main-revised}.
\end{proof}

\begin{corollary}[Recovery for a fixed post-training objective]
\label{cor:geometric-post-revised}
Assume Assumption~\ref{ass:pl-revised}.  Fix $\alpha$ and $\theta_T$ throughout these $K$ post-training steps.  If post-training uses gradient descent with
\begin{equation}
    0<\eta_{\mathrm{min}}\leq \frac{1}{L_P},
\end{equation}
then after $K$ post-training steps,
\begin{equation}
    \calP_\alpha(\theta_K)-\calP_\alpha^\star
    \leq
    \left(1-\eta_{\mathrm{min}}\mu_{PL}\right)^K
    \left(\calP_\alpha(\theta_0)-\calP_\alpha^\star\right).
    \label{eq:geometric-post-revised}
\end{equation}
Consequently, to reach optimization accuracy $\varepsilon_{\mathrm{min}}$, it suffices that
\begin{equation}
    K
    \geq
    \frac{1}{\eta_{\mathrm{min}}\mu_{PL}}
    \log\left(
    \frac{\calP_\alpha(\theta_0)-\calP_\alpha^\star}{\varepsilon_{\mathrm{min}}}
    \right).
    \label{eq:post-rounds-revised}
\end{equation}
\end{corollary}

\begin{proof}
By smoothness of $\calP_\alpha$ and the gradient-descent update,
\begin{equation}
    \calP_\alpha(\theta_{k+1})
    \leq
    \calP_\alpha(\theta_k)
    -\eta_{\mathrm{min}}\left(1-\frac{L_P\eta_{\mathrm{min}}}{2}\right)
    \|\nabla\calP_\alpha(\theta_k)\|^2.
\end{equation}
Since $\eta_{\mathrm{min}}\leq 1/L_P$,
\begin{equation}
    \calP_\alpha(\theta_{k+1})
    \leq
    \calP_\alpha(\theta_k)
    -\frac{\eta_{\mathrm{min}}}{2}\|\nabla\calP_\alpha(\theta_k)\|^2.
\end{equation}
Applying the PL condition in Equation~\eqref{eq:PL-revised} yields
\begin{equation}
    \calP_\alpha(\theta_{k+1})-\calP_\alpha^\star
    \leq
    \left(1-\eta_{\mathrm{min}}\mu_{PL}\right)
    \left(\calP_\alpha(\theta_k)-\calP_\alpha^\star\right).
\end{equation}
Iterating this recursion proves Equation~\eqref{eq:geometric-post-revised}.  Solving for $K$ gives Equation~\eqref{eq:post-rounds-revised}, using $\log(1/(1-a))\geq a$ for $a\in(0,1)$.
\end{proof}

\begin{remark}[Utility consequence]
\label{rem:utility-consequence-revised}
If the post-training phase also makes $\calL_R(\theta_U)$ small, Lemma~\ref{lem:evaluation-transfer-revised} gives the retained-data error control
\begin{equation}
    \widehat{\mathrm{err}}_R(\theta_U)
    \leq
    \frac{\calL_R(\theta_U)}{\log 2}.
\end{equation}
Thus, the same post-training mechanism that moves the model toward $\theta_R^\star$ also has a direct retained-utility interpretation. 
\end{remark}
\label{sections/appendix/communication_cost}
\end{document}